\documentclass[10pt,letterpaper]{article}

\usepackage{amsmath,amssymb}

\usepackage{changepage}

\usepackage[utf8x]{inputenc}

\usepackage{cite}

\usepackage{nameref,hyperref}

\usepackage[right]{lineno}

\usepackage[table]{xcolor}

\usepackage{array}

\newcolumntype{+}{!{\vrule width 2pt}}

\newlength\savedwidth

\newcommand\thickhline{\noalign{\global\savedwidth\arrayrulewidth\global\arrayrulewidth 2pt}%
\hline
\noalign{\global\arrayrulewidth\savedwidth}}

\usepackage{graphicx}

\begin{document}

\begin{flushleft}
{\Large
\textbf\newline{An empirical survey of data augmentation for time series classification with neural networks} 
}
\newline
\\
Brian Kenji Iwana\textsuperscript{1*},
Seiichi Uchida\textsuperscript{1}
\\
\bigskip
\textbf{1} Department of Advanced Information Technology, Kyushu University, Fukuoka, Japan
\\
\bigskip

%
%





* iwana@ait.kyushu-u.ac.jp

\end{flushleft}
\section*{Abstract}

In recent times, deep artificial neural networks have achieved many successes in pattern recognition. Part of this success can be attributed to the reliance on big data to increase generalization. However, in the field of time series recognition, many datasets are often very small. One method of addressing this problem is through the use of data augmentation. In this paper, we survey data augmentation techniques for time series and their application to time series classification with neural networks. We propose a taxonomy and outline the four families in time series data augmentation, including transformation-based methods, pattern mixing, generative models, and decomposition methods. Furthermore, we empirically evaluate 12 time series data augmentation methods on 128 time series classification datasets with six different types of neural networks. Through the results, we are able to analyze the characteristics, advantages and disadvantages, and recommendations of each data augmentation method. This survey aims to help in the selection of time series data augmentation for neural network applications.


\section{Introduction}

Time series classification attempts to categorize time series into distinct categories, and it is used for a wide range of applications. 
Some applications include the recognition of signals, biometrics, sequences, sound, trajectories, and more. 
The challenge of using time series is that time series are structural patterns that are dependent on element order. 
Note that \textit{time} does not necessarily have to represent actual time and is just used to represent the sequence order. 

Traditionally, time series classification was tackled using distance-based methods~\cite{ding2008querying}. 
However, recently, artificial neural networks have had many successes in time series classification~\cite{Schmidhuber_2015,Wang_2017}. 
For example, Recurrent Neural Networks~(RNN)~\cite{rumelhart1988learning} have had many recent successes on time series in gait recognition~\cite{Chen_2019comparative,Kluwak_2020}, biosignals~\cite{Xu_2020,Kim_2020}, and online handwriting~\cite{Sun_2016,Carbune_2020}.
Also, recent work has shown that feedforward networks such as Multi-Layer Perceptrons~(MLP) and temporal Convolutional Neural Networks~(CNN)~\cite{Lecun_1998} can also achieve competitive and sometimes better results for time series recognition~\cite{Wang_2017,bai2018empirical,Kenji_Iwana_2020}. 
Part of the recent successes of neural networks is due to the recent availability of data~\cite{Schmidhuber_2015}. 
Furthermore, it has been shown that increasing the amount of data can help with improving the generalization ability as well as the overall performance of the model~\cite{Banko_2001,Torralba_2008}. 

However, acquiring large amounts of data can be a problem for many time series recognition tasks. 
For example, the 2018 University of California Riverside~(UCR) Time Series Archive~\cite{UCRArchive2018} is one of the largest repositories of time series datasets, and out of 128 datasets, only 12 have more than a thousand training patterns. 
This demonstrates that there is a need for time series data.

One solution to acquiring more data is to generate synthetic patterns, i.e., \textit{data augmentation}. 
Notably, data augmentation is a universal model-independent data side solution. 
Data augmentation attempts to increase the generalization ability of trained models by reducing overfitting and expanding the decision boundary of the model~\cite{Shorten_2019}. 
The need for generalization is especially important for real-world data and can help networks overcome small datasets~\cite{olson2018modern} or datasets with imbalanced classes~\cite{Blagus_2013,hasibi2019augmentation}.

For image recognition, data augmentation is already an established practice. 
Most of the original proposals of the state-of-the-art Convolutional Neural Network~(CNN)~\cite{Lecun_1998} architectures used some form of data augmentation. 
For instance, AlexNet~\cite{krizhevsky2009learning}, one of the first deep CNNs that set a record benchmark on the ImageNet Large Scale Visual Recognition Challenge (ILSVRC) dataset~\cite{ILSVRC15}, used cropping, mirroring, and color augmentation. 
Other examples include the original proposal for the Visual Geometry Group~(VGG) network~\cite{simonyan2014very} which used scaling and cropping, Residual Networks~(ResNet)~\cite{He_2016} which used scaling, cropping, and color augmentation, DenseNet~\cite{Huang_2017} which used translation and mirroring, and Inception networks~\cite{Szegedy_2015} which used cropping and mirroring. 

\begin{figure}[!h]
\centering
\includegraphics{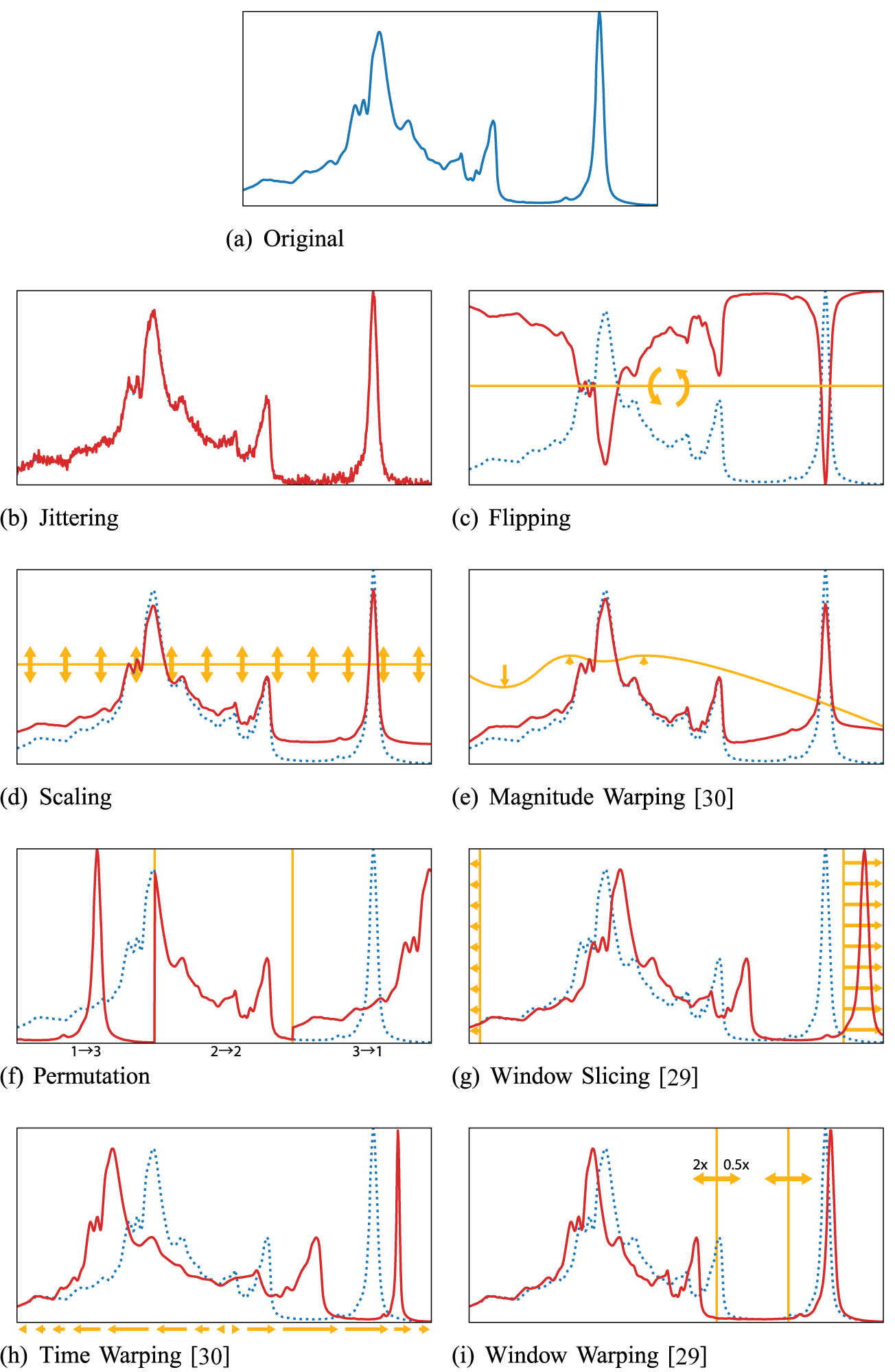}
\label{subfig:windowwarping}
\caption{{\bf Examples of random transformation-based data augmentation on the OliveOil dataset.} The dotted blue lines are the original patterns and the solid red lines are the generated patterns.}
\label{fig:random_examples}
\end{figure}

While data augmentation is a common practice in image recognition with neural networks, it is not established as a standard procedure for time series recognition~\cite{wen2020time}.
Similar to data augmentation for images, most data augmentation techniques for time series are based on random transformations of the training data. 
For example, adding random noise~\cite{Fields_2019}, slicing or cropping~\cite{le2016data}, scaling~\cite{Um_2017}, random warping in the time dimension~\cite{Um_2017,Fields_2019}, and frequency warping~\cite{jaitly2013vocal}. 
Examples of random transformation-based methods are shown in Fig.~\ref{fig:random_examples}. 
The figure shows an example pattern from the OliveOil dataset from the 2018 UCR Time Series Archive with eight random transformation-based data augmentation methods.

The problem with random transformation-based data augmentation is that there is a diverse amount of time series with each having different properties, and not every transformation is applicable to every dataset.
For example, jittering (adding noise) assumes that it is normal for the time series patterns of the particular dataset to be noisy. 
While this might be true for sensor, audio, or Electroencephalogram (EEG) data, this is not necessarily true for time series based on object contours, such as the Adiac and Fish datasets from the 2018 UCR Time Series Archive. 
These datasets are pseudo-time series taken from the contours of the objects in images. 
Another example would be domain-specific transformations, such as frequency warping for audio. 

An alternative to random transformations is to synthesize time series using information inherent to the data. 
Some examples of this are pattern mixing, generative models, and pattern decomposition methods. 
In pattern mixing, two or more existing time series are combined to produce new patterns. 
The idea is that mixing different existing patterns can create new samples with features from both patterns. 
Generative models take a less direct route and use the distributions of features in the datasets to generate new patterns. 
For example, many statistical models such as Gaussian trees~\cite{Cao_2014} and handcrafted mathematical models~\cite{Wendling_2000} have been proposed. 
Another recent generative model for time series generation is the use of neural networks, such as Generative Adversarial Networks (GAN)~\cite{goodfellow2014generative}. 
Finally, the last family is decomposition methods. 
Decomposition methods extract features from the dataset, such as trend components~\cite{Bergmeir_2016} and independent components~\cite{Eltoft}, and generate new patterns from those extracted features.
The advantage of these families of methods is that they attempt to preserve the distribution of time series in the dataset~\cite{Forestier_2017}, whereas random transformations might unintentionally change the distribution.

\begin{figure}[!h]
\centering
\includegraphics[width=1\columnwidth]{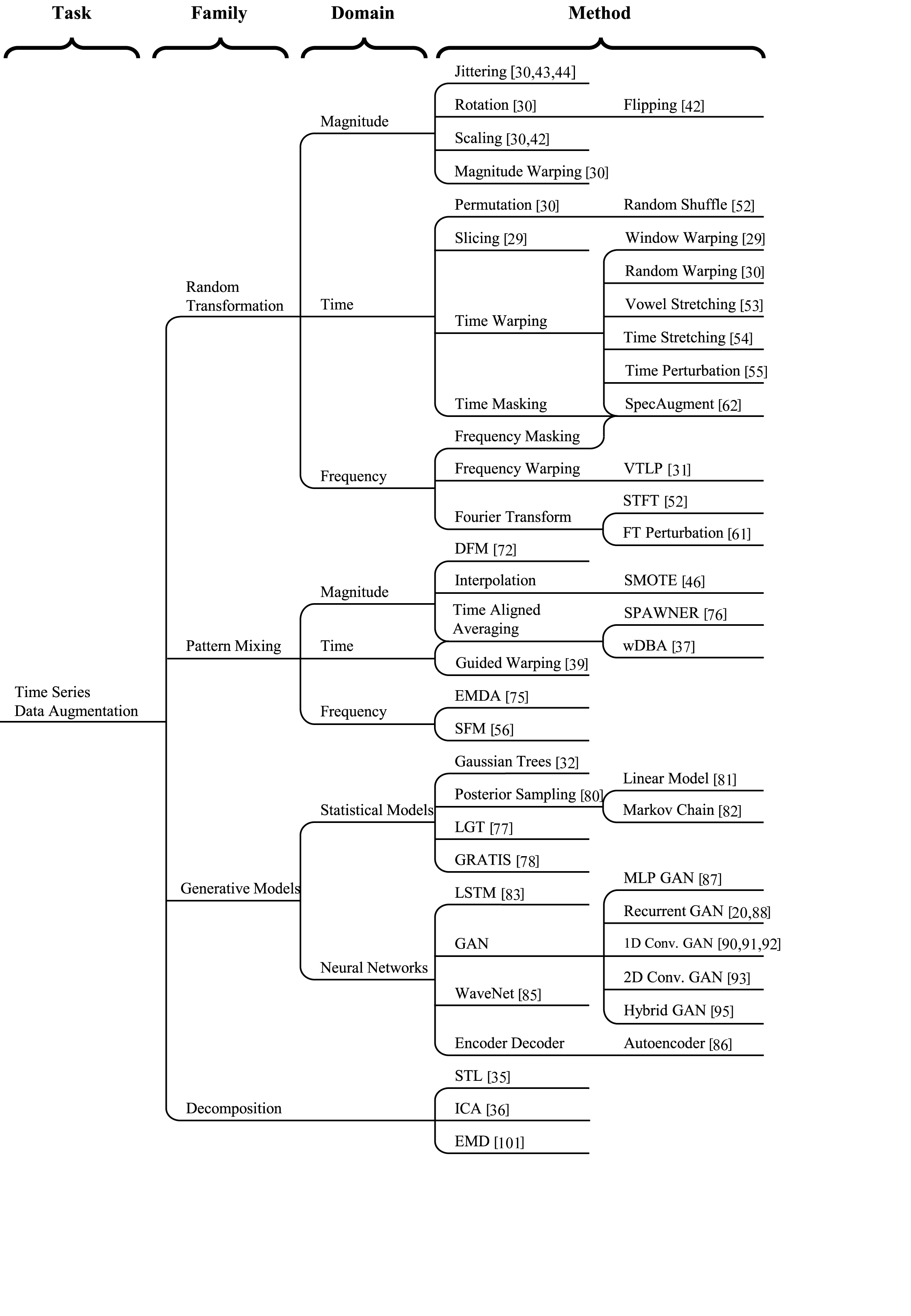}
\caption{{\bf Taxonomy of time series data augmentation.}}
\label{fig:taxonomy}
\end{figure}

A taxonomy of the time series data augmentation methods is shown in Fig.~\ref{fig:taxonomy}. 
The taxonomy breaks down time series data augmentation methods into three primary hierarchical levels, family, domain, and method.
The families of data augmentation methods include transformations, pattern mixing, generative models, and decompositions. 
The families are broken into their respective domains, or major subtypes. 

The purpose of this paper is to gather and present many data augmentation techniques for time series. 
In addition, we aim to empirically evaluate some of the techniques using a wide variety of data types. 
Data augmentation can play an important part in the pattern recognition workflow. Therefore, this is important for progress in the field.

There have been only a few works in the past that compile many data augmentation methods for time series data. 
In one example, Liu et al.~\cite{Liu_2020} compare jittering, permutation, scaling, and time warping using a Fully Convolutional Network~(FCN) and a ResNet. 
In another, Wen et al.~\cite{wen2020time} provide a high-level survey of time series augmentation methods. 
In addition to surveys, only a few works in time series recognition compare many data augmentation methods. 
These works are typically only performed on a limited amount of datasets~\cite{Um_2017} or using a limited amount of models~\cite{iwana2020time,le2016data}.
Thus, these works do not extensively evaluate the various data augmentation methods on many different datasets. 
In our work, we dive deeper into data augmentation methods for time series classification and evaluate more methods on a much larger amount of datasets with multiple neural network models.
In addition, we provide analysis, observations, and recommendations from the results, which the previous time series data augmentation survey does not.

The contributions are as follows:
\begin{itemize}
    \item We review time series data augmentation with a comprehensive taxonomy and categorization. In addition, we thoroughly outline and describe time series data augmentation methods.
    \item Using time series classification as a target, we perform a thorough comparative evaluation of time series data augmentation methods and demonstrate their effects on a variety of state-of-the-art neural network-based models. The data augmentation methods used for the evaluations include jittering, permutation, flipping, scaling, magnitude warping, time warping, slicing, window warping, SuboPtimAl Warped time series geNEratoR~(SPAWNER)~\cite{Kamycki_2019}, weighted Dynamic Time Warping Barycentric Averaging~(wDBA)~\cite{Forestier_2017}, Random Guided Warping~(RGW)~\cite{iwana2020time}, and Discriminative Guided Warping~(DGW)~\cite{iwana2020time}. Each of these methods is compared to no augmentation (or identity), using an MLP, a Long Short-Term Memory (LSTM) recurrent neural network, a Bidirectional Long Short-Term Memory (BLSTM) recurrent neural network, a VGG network, a ResNet, and a LSTM Fully Convolutional Network~(LSTM-FCN). We test on all 128 datasets from the 2018 UCR Time Series Archive repository.
    \item We discuss the aspects of each time series data augmentation method including, the advantages and disadvantages, the characteristics, and suggestions for usage for different dataset types, dataset properties, and with different models.
    \item Thorough analysis is performed using visualizations of the data augmentation methods, correlation analysis between accuracy and dataset properties, and examination of the comparative evaluations. 
\end{itemize}

The rest of this paper is structured as follows. 
Sections~\ref{sec:transformation}, \ref{sec:patternmixing}, \ref{sec:generative}, and \ref{sec:decomposition} detail transformation-based methods, pattern mixing, generative models, and decomposition methods, respectively. 
The evaluation, results, and discussion are described in Sections~\ref{sec:evaluation} and~\ref{sec:discussion}.
Finally, the conclusion and future work are provided in Section~\ref{sec:conc}.

\section{Random transformation-based data augmentation}
\label{sec:transformation}

Many earlier time series data augmentation techniques are borrowed from image data augmentation, such as cropping, flipping, and noise addition. 
These augmentation methods rely on random transformations of the training data. 
Namely, random transformation-based data augmentation generates pattern $\mathbf{x}'$ using some transformation function 
where $\mathbf{x}$ is a reference sequence $\mathbf{x}=x_1,\dots,x_t,\dots,x_T$ with $T$ number of time steps from the training set. 
Each element $x_t$ can be univariate or multivariate. 

Transformations on time series can generally be divided into three domains, the magnitude domain, time domain, and frequency domain. 
Magnitude domain transformations transform the time series along the variate or value axes. 
Time domain transformations affect the time steps and frequency domain transformations warp the frequencies. 
There are also hybrid methods that use multiple domains. 
It should be noted that multiple transformation techniques can be used to augment the data set in serial~\cite{Um_2017} and in parallel~\cite{Huang_2019,Rashid_2019}.
In the following subsections, we will detail each of these domains and the random transformation-based data augmentation methods associated with them.

\subsection{Magnitude domain transformations}

Magnitude domain transformation-based data augmentations are transformations that are performed on the values of the time series. 
The important characteristic of magnitude transformations is that only the values of each element are modified and the time steps are kept constant.

\subsubsection{Jittering}

One of the simplest, yet effective, transformation-based data augmentation methods is jittering, or the act of adding noise to time series. 
Jittering can be defined as:
\begin{equation}
    \label{eq:jittering}
    \mathbf{x}' = x_1+\epsilon_1,\dots,x_t+\epsilon_t,\dots,x_T+\epsilon_T,
\end{equation}
where $\epsilon$ is typically Gaussian noise added to each time step $t$ and $\epsilon \sim \mathcal{N}(0,\sigma^2)$. 
The standard deviation $\sigma$ of the added noise is a hyperparameter that needs to be pre-determined. 
Adding noise to the inputs is a well-known method of increasing the generalization of neural networks~\cite{Bishop_1995,An_1996}. 
It is able to do this by effectively creating new patterns with the assumption that the unseen test patterns are only different from the training patterns by a factor of noise. 
In addition, jittering has been shown to help mitigate time series drift for various neural network models~\cite{Fields_2019}. 
Time series drift when the data distribution changes due to the introduction of new data.

The use of jittering for time series has been most frequently used with sensor data. 
For example, Rashid and Louis~\cite{Rashid_2019} used a combination of jittering with other data augmentation techniques to improve the accuracy of LSTM for sensor data from construction equipment. 
Um et al.~\cite{Um_2017} also used jittering with ResNet for wearable sensor data for Parkinson's disease monitoring. 
However, the effects of jittering seem to be detrimental in the work by Um et al. 
Another example is Arslan et al.~\cite{Arslan_2019}, who used a combination of Synthetic Minority Oversampling TEchnique~(SMOTE)~\cite{Chawla_2002} and Gaussian noise for temperature, light, and air sensor data.

\subsubsection{Rotation}

Rotation is defined as:
\begin{equation}
    \label{eq:rotation}
    \mathbf{x}' = R{x}_1, \dots, R{x}_t,\dots,R{x}_T,
\end{equation}
where $R$ is an element-wise random rotation matrix for angle $\theta\sim\mathcal{N}(0,\sigma^2)$ for multivariate time series~\cite{Um_2017} and flipping for univariate time series~\cite{Rashid_2019}. 
While rotation data augmentation can create plausible patterns for image recognition, it might not be suitable for time series since rotating a time series can change the class associated with the original sample~\cite{IsmailFawaz2018}. 
This is supported by rotation augmentation being seen to have either no effect or a detrimental effect on time series classification with neural networks~\cite{Ohashi_2017,Rashid_2019,iwana2020time}. 
Conversely, Um et al.~\cite{Um_2017} found that rotation data augmentation did improve accuracy, especially when combined with other augmentation methods.

\subsubsection{Scaling}

Scaling changes the global magnitude, or intensity, of a time series by a random scalar value. 
With scaling parameter $\alpha$, scaling is a multiplication of $\alpha$ to the entire time series, or:
\begin{equation}
    \label{eq:scaling}
    \mathbf{x}' = {\alpha}{x}_1, \dots, {\alpha}{x}_t,\dots,{\alpha}{x}_T,
\end{equation}
The scaling parameter $\alpha$ can be determined by a Gaussian distribution $\alpha\sim\mathcal{N}(1,\sigma^2)$ with $\sigma$ as a hyperparameter~\cite{Um_2017}, or it can be from a random value from a pre-defined set~\cite{Rashid_2019}.
It should be noted that ``scaling'' in terms of time series is different than in the image domain. 
For time series, it refers to just increasing the magnitude of the elements and not enlarging the time series.
Some examples of using scaling as data augmentation include classification of sensor data~\cite{Um_2017,Rashid_2019}. 
Escano et al.~\cite{Delgado_Escano_2019} and Tran and Choi~\cite{Tran_2020} used a combination of scaling with jittering and element-wise interpolation for Gait recognition.

\subsubsection{Magnitude warping}

Magnitude warping~\cite{Um_2017} is a time series specific data augmentation technique that warps a signal's magnitude by a smoothed curve. 
Namely, augmented time series $\mathbf{x}'$ is:
\begin{equation}
    \label{eq:magwarp}
    \mathbf{x}' = {\alpha}_1{x}_1, \dots, {\alpha}_t{x}_t,\dots,{\alpha}_T{x}_T,
\end{equation}
where ${\alpha}_1,\dots,{\alpha}_t,\dots,{\alpha}_T$ is a sequence created by interpolating a cubic spline $S(\mathbf{u})$ with knots $\mathbf{u}=u_1,\dots,{u}_i,\dots,{u}_I$. 
Each knot $u_i$ is taken from a distribution $\mathcal{N}(1,\sigma^2)$ where the number of knots $I$ and the standard deviation $\sigma$ are hyperparameters. 
The idea behind magnitude warping is that small fluctuations in the data can be added by increasing or decreasing random regions in the time series. 
However, the downsides of magnitude warping for data augmentation is that it still assumes the random transformation is realistic and it depends on two pre-defined hyperparameters (the number of knots $I$ and the standard deviation of the knot height $\sigma$) instead of one like many of the other transformation-based methods.

\subsection{Time domain transformations}

Time domain transformations are similar to magnitude domain transformations except that the transformation happens on the time axis. 
In other words, the elements of the time series are displaced to different time steps than the original sequence.
The following methods are common examples of time domain transformations.

\subsubsection{Slicing}

Slicing is the time series data augmentation equivalent to cropping for image data augmentation. 
The general concept behind slicing is that the data is augmented by slicing time steps off the ends of the pattern, or:
\begin{equation}
    \label{eq:window_warp}
    \mathbf{x}'=x_{\varphi},\dots,x_{t},\dots,x_{W+\varphi}
\end{equation}
where $W$ is the size of a window and $\varphi$ is a random integer such that $1\leq \varphi \leq T-W$.
Slicing in this way is also sometimes referred to as Window Slicing~(WS)~\cite{le2016data} due to the use of a window of size $W$.

\subsubsection{Permutation}

Permutation for data augmentation was proposed by Um et al.~\cite{Um_2017} as a method of rearranging segments of a time series in order to produce a new pattern. 
It should be noted that permutation does not preserve time dependencies.
It can be performed in two ways, with equal sized segments and with variable sized segments~\cite{Pan_2020}. 
Using permutation with equal sized segments splits the time series into $N$ number of segments of length $\frac{T}{N}$ and permutes them.
Using variable size segments uses segments of random sizes.

Random shuffling can be considered as a form of permutation which rearranges individual elements rather than segments. 
In one example of random shuffling, Eyobu and Han~\cite{Steven_Eyobu_2018} incorporated a shuffling step into their data augmentation workflow of feature extraction, local averaging, shuffling, and local averaging again for sensor data classification with LSTMs.

\subsubsection{Time warping}
\label{subsec:timewarping}

Time warping is the act of perturbing a pattern in the temporal dimension. 
This can be performed using a smooth warping path~\cite{Um_2017} or through a randomly located fixed window~\cite{le2016data}.
When using time warping with a smooth warping path, the augmented time series becomes:
\begin{equation}
    \label{eq:time_warp}
    \mathbf{x}'=x_{\tau(1)},\dots,x_{\tau(t)},\dots,x_{\tau(T)},
\end{equation}
where $\tau( \cdot )$ is a warping function that warps the time steps based on a smooth curve. 
The smooth curve is defined by a cubic spline $S(\mathbf{u})$ with knots $\mathbf{u}=u_1,\dots,{u}_i,\dots,{u}_I$.
The height of the knots $u_i$ taken from $u_i\sim\mathcal{N}(1,\sigma^2)$. 
In this way, the time steps of the series have a smooth transition between stretches and contractions. 

Alternatively, a popular method of time warping called window warping was proposed by Le Guennec et al.~\cite{le2016data}. 
Window warping takes a random window of the time series and stretches it by 2 or contracts it by $\frac{1}{2}$. 
While the multipliers are fixed to $\frac{1}{2}$ and 2, Le Guennec et al. note that they can be modified or optimized to other values. 

Similarly, independently developed ideas are vowel stretching~\cite{Nagano2019}, dynamic time stretching~\cite{Nguyen_2020}, and time perturbation~\cite{Vachhani_2018} for speech data augmentation. 
Vowel stretching targets vowels and extends them by interpolating frames.
Time perturbation re-samples the input signal by a randomly selected factor.

\subsection{Frequency domain transformations}

Frequency domain transformations are transformations that are specific to periodic signals, such as acoustic data. 
The following are common methods of frequency domain transformations for data augmentation.

\subsubsection{Frequency warping}

In audio and speech recognition, frequency warping is a popular method of data augmentation~\cite{jaitly2013vocal,Cui_2014}. 
Vocal Tract Length Perturbation (VTLP)~\cite{jaitly2013vocal}, for example, is an extension to Vocal Tract Length Normalization (VTLN)~\cite{Lee_1998} that adds variability instead of removing it. 
In VTLP, frequency $f$ is mapped to a new frequency $f'$ using:
\begin{equation}
    \label{eq:frequency_warp}
    f' = f\omega,
\end{equation}
if $f\leq F_{\mathrm{hi}} \frac{\min(\omega,1)}{\omega}$ and:
\begin{equation}
    f'=\frac{s}{2}-\frac{\frac{s}{2}-F_{\mathrm{hi}} \min(\omega,1)}{\frac{s}{2}-F_{\mathrm{hi}} \frac{\min(\omega,1)}{\omega}},
\end{equation}
otherwise, 
where $\omega$ is a random warp factor, $s$ is the sampling frequency, and $F_{\mathrm{hi}}$ is a boundary frequency.
The warping is applied directly to the Mel filter banks.
VTLP is a popular data augmentation method and has been used for many audio applications, such as vocal tract shape conversion~\cite{Adachi_2007} and acoustic modeling~\cite{Cui_2014,ko2015audio}.
It has also been extended by using it in an end-to-end recognition framework~\cite{Kim_2019}.

\subsubsection{Fourier transform-based methods}

There have also been data augmentation methods that augment by manipulating the data under a Fourier transform. 
Gao et al.~\cite{gao2020robusttad} proposed utilizing amplitude and phase perturbations in order to augment in the frequency domain. 
This is done by adding Gaussian noise to the amplitude and phase spectra found by a discrete Fourier transform. 
In another example, Eyobu and Han~\cite{Steven_Eyobu_2018} use Short-Time Fourier transform (STFT) features as one of the features augmented using their method. 




\subsubsection{Spectrogram augmentation}

Normally, frequency warping data augmentation is performed before conversion into a spectrogram. 
However, recently, a method called SpecAugment~\cite{Park_2019} was proposed that augments the spectrogram data itself. 
In order to augment the data, SpecAugment performs three key operations on the spectrogram: time warping, frequency masking, and time masking. 
SpecAugment's time warping works much like the window warping method, except with random duration. 
Frequency masking and time masking mask the spectrograms in their respective domains.
In this way, SpecAugment is both a time domain and frequency domain augmentation method.

\section{Pattern mixing}
\label{sec:patternmixing}

Pattern mixing combines one or more patterns to generate new ones. 
For random transformations, there is an assumption that the results of the transformations are typical of the dataset. 
However, not every transformation is applicable to every dataset. 
The benefit of pattern mixing is that it does not make this same assumption.
Instead, pattern mixing assumes that similar patterns can be combined and have reasonable results. 

\subsection{Magnitude domain mixing}

The most direct application of pattern mixing is to linearly combine the patterns at each time step. 
This is the idea behind magnitude domain mixing. 

\subsubsection{Averaging and interpolation}

It is possible to create new patterns by simply averaging two patterns. 
In general, the reference patterns are selected randomly from the same class or selected using nearest neighbors. 
Interpolation extends averaging to more points between the two patterns instead of just the midpoint.

One famous interpolation method is called SMOTE~\cite{Chawla_2002}. 
SMOTE was designed to combat data with imbalanced classes by interpolating patterns from under-represented classes. 
In SMOTE, a random sample $\mathbf{x}$ is selected from the under-represented class and another random sample $\mathbf{x}_{\mathrm{NN}}$ is selected from the reference sample's $k$-nearest neighbors.
Next, the difference between the two samples is multiplied by a random value $\lambda$ in a range of $\{0, 1\}$. 
The result is a pattern between the two original patterns, or:
\begin{equation}
    \label{eq:smote}
    \mathbf{x}'=\mathbf{x}+\lambda |\mathbf{x}-\mathbf{x}_{\mathrm{NN}}|.
\end{equation}
SMOTE has been shown to perform well in many time series applications, such as sensor data~\cite{Arslan_2019}, gene sequences~\cite{Khadijah_2018}, high-dimensional data~\cite{Blagus_2013}. 
There have also been a number of improvements on SMOTE, such as Safe-Level-SMOTE~\cite{Bunkhumpornpat_2009}, SMOTE based on the furthest neighbor~(SMOTEFUNA)~\cite{Tarawneh_2020}, Cost Minimization Oriented SMOTE~(CMO-SMOTE)~\cite{Zhou_2016},  Density-Based SMOTE~(DBSMOTE)~\cite{Bunkhumpornpat_2011}, etc.
SMOTE has also been used in a feature space modeled by an Echo state network (ESN)~\cite{Gong_2016}.

In addition to SMOTE, there have been other proposals of interpolation for data augmentation. 
For example, Sawicki and Zielinski~\cite{Sawicki_2020} used interpolation in combination with LSTMs on sensor data.
In addition, an interpolation method similar to SMOTE was extended by DeVries and Taylor~\cite{devries2017dataset} to extrapolation by allowing $\lambda$ in Eq.~\eqref{eq:smote} to be $\{0, \infty\}$. 

\subsubsection{Deviation from the mean}

Yeomans et al.~\cite{Yeomans_2019} proposed a method of time series data augmentation using the deviation from the mean~(DFM). 
To simulate new time series, they use the following process. 
First, the signals are smoothed with a Savitzky-Golay filter~\cite{Savitzky_1964} and an offset is used to ensure that all values are greater than 0. 
Next, the bounding curves of the smoothed signals for each class are calculated. 
A mean curve is then calculated using the bounding curves and the DFM for each pattern is determined based on the difference between the pattern and the mean curve of its class. 
Random segments of DFMs from multiple patterns are then combined to create a surrogate DFM curve. 
Finally, the surrogate DFM is multiplied with the class mean curve to create new simulated patterns.

\subsection{Time domain mixing}

Guided warping~\cite{iwana2020time} combines time series by time warping a reference pattern by a teacher pattern using Dynamic Time Warping (DTW)~\cite{sakoe1978dynamic}. 
DTW is a method of measuring the distance between time series using elastic element matching found by dynamic programming. 
Guided warping uses the dynamic alignment function of DTW to warp the elements of a reference pattern to the elements of a teacher pattern. 
In this way, the reference pattern is set to the time steps of the teacher pattern. 
This is different from averaging in that the mixing happens only in the time domain.
There are two variants, Random Guided Warping (RGW) which uses a random intra-class teacher and Discriminative Guided Warping (RGW) which uses a directed discriminative teacher~\cite{iwana2020time}.

\subsection{Frequency domain mixing}

Pattern mixing can also be performed in the frequency domain. 
Takahashi et al.~\cite{Takahashi_2016} proposed a method called Equalized Mixture Data Augmentation~(EMDA), which mixes two sounds of the same class with randomly selected timings. 
In addition to mixing sounds, EMDA perturbs the sound by boosting or attenuating particular frequency bands. 
In another example, Stochastic Feature Mapping (SFM)~\cite{Cui_2014} converts one speaker's speech data to another speaker by mapping features using an acoustic model. 
Due to the nature of using the frequency domain, data augmentations in this area are generally used for sound recognition, e.g., EMDA has been used for acoustic event detection~\cite{Takahashi_2016} and animal audio classification~\cite{Nanni_2020} and SFM has been used for speech~\cite{Cui_2014}.

\subsection{Mixing in multiple domains}

There are also methods that use pattern mixing across multiple domains. 
For instance, the following methods mix the patterns in multiple domains.

\subsubsection{Suboptimal element alignment averaging}

SuboPtimAl Warped time series geNEratoR (SPAWNER)~\cite{Kamycki_2019} was introduced by Kamycki et al. as a method of generating patterns through a novel method called suboptimal time warping. 
Suboptimal time warping uses the warping ability of DTW but adding an additional constraint that forces the warping path through a random point. 
By using the suboptimal time warping, SPAWNER is able to create an almost unlimited number of new time series by averaging aligned patterns.

\subsubsection{Barycentric averaging}

DTW Barycentric Averaging~(DBA)~\cite{Petitjean_2011} is a method of averaging multiple discrete time series by finding the center of time aligned elements. 
It does this through an iterative process of adding sample patterns that are time aligned by DTW to a cumulative centroid pattern. 
The advantage of using DBA over linear averaging is that the underlying pattern is preserved, whereas linear averaging might smooth features (for example, linear averaging time series that are just offset in time would lose distinct features).

For data augmentation, Forestier et al.~\cite{Forestier_2017} proposed a weighted version of DBA~(wDBA). 
They propose three weighting schemes, Average All (AA), Average Selected (AS), and Average Selected with Distance (ASD). 
AA weights all of the time series in a class input into wDBA by a flat Dirichlet distribution. 
AS selects a reference time series and weights two of the five nearest neighbors by a large constant amount and all the rest by a small constant amount. 
ASD is similar to AS except that it weights based on the distance to the reference. 
Fawaz et al.~\cite{IsmailFawaz2018} also used wDBA AS with a ResNet for time series classification.

\section{Generative models}
\label{sec:generative}

Instead of using random transformations or mixing patterns, it is possible to sample time series from feature distributions using generative models. 
We classify generative models into two categories, statistical models and neural network-based models.

\subsection{Statistical generative models}

There are a wide variety of statistical, mathematical, or stochastical models used for time series generation and augmentation. 
Typically, these augmentation methods build a statistical model of the data and are often used in forecasting. 
For example, the Local and Global Trend (LGT)~\cite{smyl2016data} is a time series forecasting model that uses nonlinear global trends and reduced local linear trends to model the data. 
LGT-based data augmentation has been shown to improve forecasting results with LSTMs~\cite{smyl2016data}. 
In Cao et al.~\cite{Cao_2014}, a mixture of Gaussian trees was used to oversample imbalanced classes for time series classification. 
GeneRAting TIme Series (GRATIS)~\cite{Kang_2020} was recently introduced, and it uses mixture autoregressive (MAR) models in order to simulate time series. 
GRATIS can be used to generate non-Gaussian and nonlinear time series by modeling with MAR and adjusting the parameters.
There are also works that use the posterior sampling technique proposed by Tanner and Wong~\cite{Tanner_1987}, such as a dynamic linear model with a hyperparameter approximated from marginal probability functions~\cite{Fr_hwirth_Schnatter_1994} and posterior sampling a Markov chain Monte Carlo~\cite{Meng_1999} model. 

\subsection{Neural network-based generative models}

Generative models based on neural networks have become popular in recent times. 
However, while many generative networks have been proposed, not every model was used for data augmentation. 
In this section, we will discuss the networks that have specifically been used for data augmentation.


The most basic application of neural networks for time series generation is direct sequence-to-sequence networks such as LSTMs and temporal CNNs.
This technique is especially useful for natural language processing~(NLP) tasks. 
For example, Hou et al.~\cite{hou2018sequence} tackle language understanding and demonstrate the effectiveness of an LSTM-based input-feeding neural machine translation~(NMT) model with attention in generating sentences for data augmentation.
Longpre et al.~\cite{Longpre_2019} use a back-translation data augmentation strategy with sequence-to-sequence networks for the question-answer task. 
Another example is the use of WaveNet~\cite{oord2016wavenet}, a speech generation network that uses dilated causal convolutions, which has been used for data augmentation~\cite{Wang_2019wave}.

\subsubsection{Encoder-decoder networks}

Encoder-decoder networks take a high-dimensional or structural input, encode it into a latent space lower-dimensional vector, and then decode it back to a high-dimensional or structural output. 
Data augmentation methods generate new patterns by decoding vectors sampled from the latent space. 
In one example, an LSTM-based autoencoder (LSTM-AE)~\cite{Tu_2018} was used to generate data for a classification LSTM on skeleton-based human action recognition. 
However, the results were mixed when comparing the results of data augmentation using the LSTM-AE, rotation, scaling, and no augmentation. 
On one dataset, the data augmentation from LSTM-AE had the best results, but on the two others, no augmentation was better.
In another example, DeVries and Taylor~\cite{devries2017dataset} combined samples generated from an LSTM-based variational autoencoders~(VAE) and combined it with interpolation and extrapolation to augment time series.

\subsubsection{Generative adversarial networks}

GANs~\cite{goodfellow2014generative} are a class of generative networks that use adversarial training to jointly optimize two neural networks, a generator and a discriminator. 
Similar to encoder-decoder networks, in order to generate samples, the GAN is trained using the training dataset and then $z$-vector is sampled and used with the generator to create new time series.
There have been numerous time series GANs proposed. However, most target generation only and not data augmentation. 
Due to this, we will only focus on the works that are specifically used for data augmentation. 

The underlying networks of GANs for time series can be roughly separated into four architectures, GANs based on fully-connected networks or MLPs, recurrent GANs that use RNNs, GANs with temporal CNNs or 1D CNNs, and GANs that generate spectrum based images with 2D CNNs. 
Lou et al.~\cite{Lou_2018} is an example of a GAN built on a fully-connected network. 
They combine an autoencoder network with a Wasserstein GAN~(WGAN) in order to augment time series regression data.

Some examples of recurrent GANs include~\cite{Haradal_2018} and~\cite{hasibi2019augmentation}. 
In the former, Harada et al.~\cite{Haradal_2018} use a straightforward implementation of a deep LSTM-based GAN for the data augmentation of ECG and EEG signals. 
They found an improvement in accuracy when compared to noise addition, interpolation, and generation with a Hidden Markov Model~(HMM). 
Similar results were found in \cite{hasibi2019augmentation} for recognizing network traffic and \cite{esteban2017real} for synthetic and medical time series.

Temporal Convolutional GANs~(T-CGAN)~\cite{ramponi2018t} are GANs that use 1D convolutional layers for time series generation.
Electronic Health Records GAN (ehrGAN)~\cite{Che_2017} is another 1D convolutional GAN that differs from a T-CGAN in that it incorporates an encoder with variational contrastive divergence in the generator.
Chen et al.~\cite{Chen_2019} proposed EmotionalGAN which is also based on 1D CNNs for classifying emotions from long ECG patterns. 
They found significant improvements when augmenting data for Random Forests and Support Vector Machines~(SVM). 
However, Hatamian et al.~\cite{Hatamian_2020} had mixed results when comparing a 1D convolutional GAN to a Gaussian Mixture Model~(GMM) on ECG data. 
There are also spectrogram-based augmentation methods there use 2D convolutional layers in their GAN, such as WaveGAN~\cite{Madhu_2019}.

Finally, there are hybrid GAN models, such as the BLSTM-CNN GAN proposed by Zhu et al.~\cite{Zhu_2019}. 
In Zhu et al., they found that their hybrid BLSTM-CNN performed better than other LSTM-based GANs.

Conditional GANs~(cGAN)~\cite{mirza2014conditional} have also been used for data augmentation. 
cGANs are GANs that are provided a condition, or parameter, to the generator and discriminator in order to control the generated patterns.
Nikolaidis et al.~\cite{nikolaidis2019augmenting} showed modest improvements on ECG data using a recurrent cGAN with an MLP, random forest, $k$-nearest neighbor, and SVM. 
Similar results were found in~\cite{Harada_2019}.
However, it is not clear if the use of cGANs is better for data augmentation because Sheng et al.~\cite{Sheng_2019} compared a traditional GAN and a cGAN on speech recognition and found that the traditional GAN performed better on average. 
Wang et al.~\cite{Wang_2018} also proposed a selective WGAN (sWGAN) and selective VAE (sVAE) that showed better performance than a standard conditional WGAN~(cWGAN).

\section{Time series decomposition}
\label{sec:decomposition}

Decomposition methods generally decompose time series signals by extracting features or underlying patterns. 
These features can either be used independently, recombined, or perturbed for generating new data for augmentation. 
Empirical Mode Decomposition (EMD)~\cite{Huang_1998} is a method of decomposing nonlinear and non-stationary signals. EMD has shown to improve classification by using it as a decomposition method for data augmentation of noisy automobile sensor data in a CNN-LSTM~\cite{Nam_2020}.
Another example of a decomposition method used for data augmentation was proposed in~\cite{Eltoft}, where the use of Independent Component Analysis~(ICA)~\cite{Comon_1994} was combined with a Dynamical-Functional Artificial Neural Network~(D-FANN) for filling gaps in time series. 
This work assumes that the observed signals are generated from independent sources and estimates the mixture using ICA. 
Using the transformed space from ICA, D-FANN is used for each component and then transformed back into the signal. 
Using this technique, Eltoft was able to increase the performance of an MLP.

There have also been methods that used Seasonal and Trend decomposition using Loess~(STL)~\cite{cleveland1990stl}. 
STL is traditionally used to decompose signals into seasonal, trend, and remainder components. 
Bergmeir et al.~\cite{Bergmeir_2016} exploited STL by decomposing the signal into these components and bootstrapping the remainder using a moving block bootstrap. 
They then assembled a new signal using the bootstrapped remainder.
Robust Time series Anomaly Detection (RobustTAD)~\cite{gao2020robusttad} also used a version of STL, namely RobustSTL~\cite{Wen_2019}. 
In RobustTAD, the signals are decomposed and augmented using a variety of time and frequency domain transformations in order to augment data for anomaly detection. 

\section{Comparative evaluations}
\label{sec:evaluation}

In this section, we describe the comparative evaluations performed using the data augmentation methods. 
The purpose of the evaluations is to empirically compare data augmentation methods on a variety of neural network models and time series datasets. 

\subsection{Datasets}

We used all of the datasets in the 2018 UCR Time Series Archive~\cite{UCRArchive2018}. 
Information about each of the datasets is outlined in \nameref{S1_Table}.
In total, 128 datasets are used, including 9 device, 6 ECG, 2 electrooculography (EOG),  2 electrical penetration graph (EPG), 3 hemodynamics (Hemo), 1 High-Resolution Melting (HRM), 32 object contours from images, 17 motion, 1 power, 30 sensor, 8 simulated, 8 spectro, 4 spectrum, 2 traffic, and 3 trajectory time series. 
The datasets have fixed training sets and test sets.

The time series are rescaled so that the smallest value in the training set is -1 and the largest value in the training set is 1. 
In addition, datasets that vary in length are zero-padded for batch training and missing values are replaced with zeroes.

\subsection{Evaluated network models}

\begin{figure}[!h]
\centering
\includegraphics{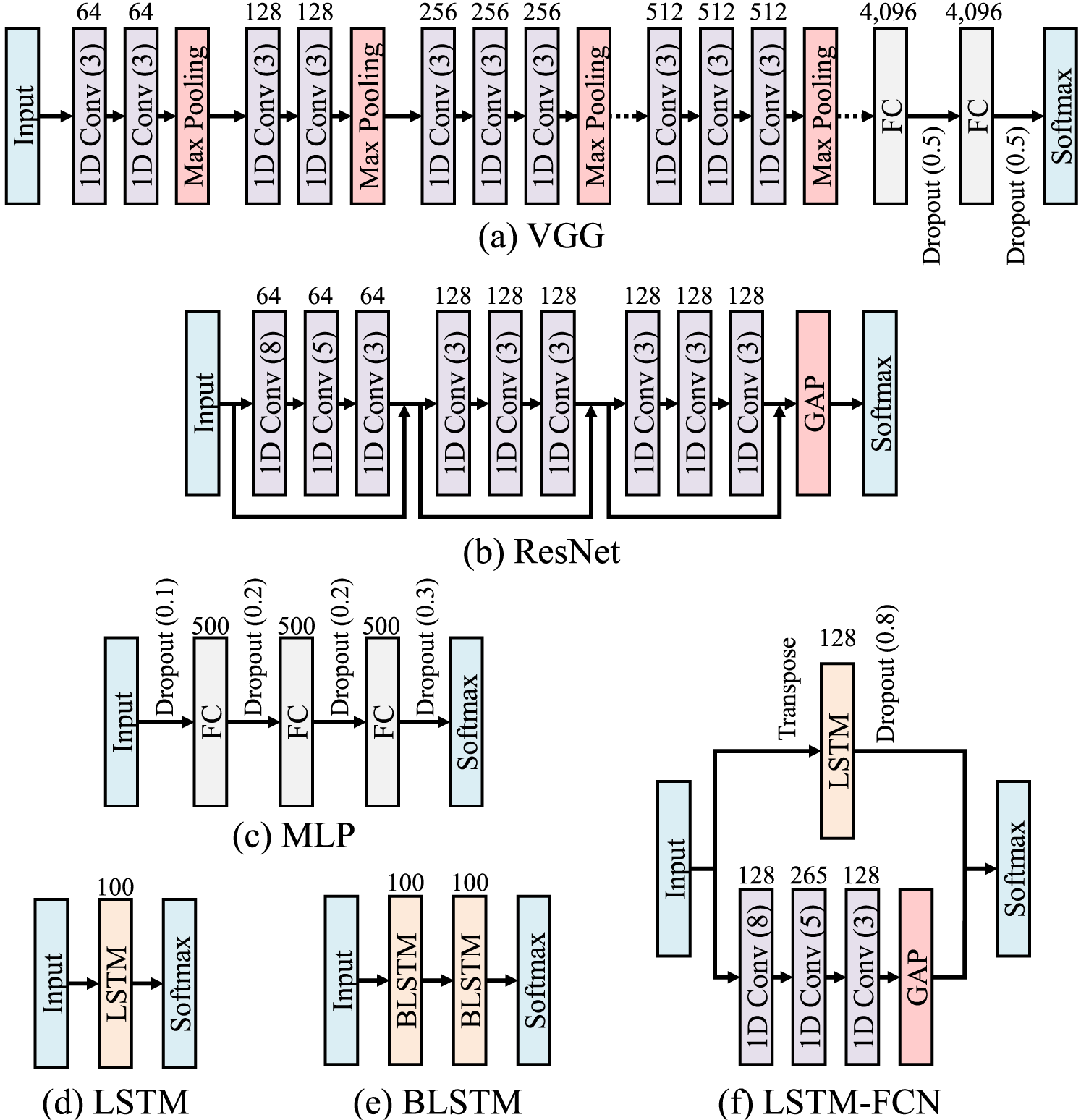}
\caption{{\bf The architectures of the neural networks used to evaluate the data augmentation methods.}}
\label{fig:networks}
\end{figure}

In order to evaluate the data augmentation methods, we used a variety of neural network models. 
This includes a 1D VGG, 1D ResNet, MLP, LSTM, BLSTM, and LSTM-FCN, as shown in Fig.~\ref{fig:networks}. 
These networks were chosen due to being state-of-the-art in time series recognition as well as representing a wide range of models with different attributes.

Each evaluation used hyperparameters taken from state-of-the-art time series classification models found in their respective literature. 
In addition, each evaluation was trained with their recommended optimizer, learning rate, and batch size. 
The only exception is that they all were trained for the same 10,000 iterations, had a learning rate reduction of 0.1 on plateaus of 500 iterations, and that the training dataset was augmented with four times the number of patterns of the original dataset. The number of iterations was kept constant with no early stopping across all of the datasets in order to give each model the same opportunity for training so that intra-model comparison between data augmentation methods is more consistent. 
A total of 10,000 iterations was chosen because the training loss for all of the models on each of the datasets tended to converge before then. 
Also, since the number of iterations is kept constant, the control experiment with no augmentation is the same as repeating identical patterns to match the augmented experiments.
In addition, it should be noted that the hyperparameters of the models have little tuning due to the inter-model accuracy being not as important as the intra-model accuracy for the purposes of data augmentation comparison. 
However, the hyperparameters used in the experiments were chosen based on a good faith attempt to find the best hyperparameters found in literature. 
The details of the hyperparameters and the training scheme are described as follows.

\subsubsection{1D Visual Geometry Group (VGG) network}

Temporal CNNs are CNNs~\cite{Lecun_1998} that use 1D convolutions instead of 2D convolutions that are traditionally used for image recognition. 
Thus, we adapted a VGG~\cite{simonyan2014very} for time series through the use of 1D convolutions.
We use the hyperparameters of the original VGG as shown in Fig.~\ref{fig:networks}~(a). 
However, because the datasets in the 2018 UCR Time Series Archive come in a wide variety of sequence sizes, the standard VGG is not appropriate due to the possibility of excessive pooling. 
Thus, we use a modified version that contains different numbers of convolutional and max pooling blocks depending on the input length~\cite{iwana2020time}.
The number of blocks $B$ used is:
\begin{equation}
    \label{eq:numpool}
    B = \textrm{round}(\log_2(T))-3,
\end{equation}
where $T$ is the number of time steps. 
This keeps the output of the final max pooling to be between 5 and 12 time steps~\cite{Kenji_Iwana_2020}. 
Similar to the original VGG, the first two blocks of convolutional layers have two consecutive convolutional layers of 64 and 128 filters, respectively. 
Every block thereafter contains three consecutive convolutional layers. 
256 filters are used for the third block and every subsequent block contains 512 filters. 
Each block is followed by max pooling with filter size 2 at stride 2. 
The convolutional layers all use ReLU activation functions, filter length of 3 at stride 1, and initialized using the uniform variance proposed by He et al.~\cite{He_2015_ICCV}.
Furthermore, after the final max pooling layer, there are two fully-connected hidden layers with 4,096 nodes each, ReLU, and dropout with a probability of 0.5. 
For training, the VGG evaluation was trained using batch size 256 with Stochastic Gradient Descent~(SGD) with a learning rate of 0.01, weight decay of $5\times 10^{-4}$, and momentum of 0.9. 

\subsubsection{1D Residual Network (ResNet)}

A ResNet is a deep CNN that uses residual connections between blocks~\cite{He_2016}.
The hyperparameters for the 1D ResNet used in the evaluation were taken from Fawaz et al.~\cite{IsmailFawaz2018}, who saw improvements using data augmentation for time series. 
As shown in Fig.~\ref{fig:networks}~(b), this version deviates from the original proposal of ResNet~\cite{He_2016} by containing only three residual blocks with varying filter lengths and no max pooling.
The first block uses three layers of 64 1D convolutions, the second block has three layers of 128 1D convolutions, and the third block has three layers of 128 convolutions. 
Each residual block contains three convolutional layers with filter size 8, 5, and 3.
Each convolution is followed by batch normalization~\cite{batchnorm} and ReLU activation functions. 
In addition, the residual connection connects the input of each residual block to the input of the next block using an addition operation. 
The last two layers of the ResNet include a Global Average Pooling~(GAP) layer and an output layer with softmax. 
All of the network parameters for the ResNet were initialized using Glorot's Uniform initialization~\cite{glorot2010understanding}. 
Finally, the network was trained using Adam optimizer~\cite{kingma2014adam} with the initial learning rate set to 0.001 and the exponential decay rates of the first and second momentum estimates set to 0.9 and 0.999 respectively, as per~\cite{IsmailFawaz2018}. 

\subsubsection{Multi Layer Perceptron (MLP)}

For this evaluation, we used a fully-connected MLP network. 
While not traditionally used for time series, MLPs have shown~\cite{Wang_2017} to be as effective as time series specific models. 
This network was selected as an evaluated model to represent a neural network that does not take structural relationships into consideration. 
In order to use an MLP for time series, the time series is flattened so that each time step is one element in the input vector. 
The version of MLP that was used was the MLP proposed by Wang et al.~\cite{Wang_2017} and is shown in Fig.~\ref{fig:networks}~(c). 
The network is constructed of three hidden layers with 500 nodes each, rectified linear unit~(ReLU) activations, and an output layer with softmax. 
Dropout is used between each layer with a rate of 0.1 after the input layer, 0.2 between the hidden layers, and 0.3 before the output layer.
As suggested by Wang et al., the MLP is optimized using Adadelta~\cite{zeiler2012adadelta} with a learning rate of 0.1, $\rho=0.95$, and $\epsilon=10^{-8}$. 
All datasets and data augmentation methods were trained with a batch size of 256.

\subsubsection{Long Short-Term Memory (LSTM)}

RNN-based models use memory states with recurrent connections to address the difficulties with time series recognition. 
Thus, using RNN-based models is useful to assess the effects of the data augmentation techniques on time distortion invariant models. 
LSTMs~\cite{Hochreiter_1997} are one of the most commonly used RNNs. 
The hyperparameters selected for the experiments were determined by the LSTM hyperparameter survey in~\cite{reimers2017optimal}. 
Namely, as shown in Fig.~\ref{fig:networks}~(d), the LSTM has one LSTM layer with 100 units. 
Both models were trained using Nesterov Momentum Adam~(Nadam)~\cite{dozat2016incorporating} optimizers with an initial learning rate of 0.001 and batch size of 32. 

\subsubsection{Bidirectional Long Short-Term Memory (BLSTM)}

BLSTMs~\cite{Schuster_1997} are a bidirectional variant of LSTMs that use a forward and backward recurrent connection. 
The idea of using both a standard LSTM and a bidirectional one is to observe the differences that having forward and backward recurrent connections has on different data augmentation methods. 
The hyperparameters and training of the BLSTM are identical to the LSTM, except that the BLSTM had two layers.
In addition, the BLSTM uses concatenation as the output merging method.

\subsubsection{Long Short-Term Memory Fully Convolutional Network (LSTM-FCN)}

The final model for the data augmentation evaluations is using the hybrid of LSTM and CNN referred to as an LSTM-FCN~\cite{Karim_2018}. 
An LSTM-FCN is a two-stream network that combines a fully convolutional stream and a recurrent stream. 
The LSTM-FCN and hyperparameters are shown in Fig.~\ref{fig:networks}~(f).
The convolutional stream has three 1D convolutional layers, each with batch normalization and ReLU. 
The convolutional filters are initialized using He's Uniform. 
For the recurrent stream, there is one LSTM layer with 128 units and an aggressive dropout rate of 0.8. 
It should be noted that the input of the LSTM layer is transposed so that the LSTM receives a $T$ dimensional vector that is one time step long. 
While this is a non-standard use of LSTM, it has shown to be more effective than the standard use~\cite{Karim_2019}. 
The output of the LSTM is concatenated with the output of GAP from the convolutional stream. 
For training, as suggested, Adam optimizer is used with batches of 128 and an initial learning rate of 0.001.

\subsection{Evaluated data augmentation methods}

The following time series data augmentation methods are evaluated.
These data augmentation methods were selected as the methods that fell under two criteria. 
First, the evaluated data augmentation methods are general methods that can be used with any time series. 
For example, we do not use any frequency domain methods due to them not being applicable to non-periodic time series. 
Second, we did not select data augmentation methods that required external training, such as the generative models.

\begin{itemize}
    \item \textbf{None}: ``None'' refers to no augmentation, which is the control experiment against which the data augmentation methods can be compared.
    \item \textbf{Jittering}: For Jittering, Gaussian noise with a mean $\mu=0$ and standard deviation $\sigma=0.03$ is added to the time series, as suggested by Um et al.~\cite{Um_2017}
    \item \textbf{Rotation}: Because the 2018 UCR Time Series Archive contains univariate time series, we use flipping as the Rotation data augmentation method. To do so, 50\% of the patterns of the training set are inverted at random for each data augmentation set~\cite{Rashid_2019}.
    \item \textbf{Scaling}: Scaling multiplies training set time series with random scalars from a Gaussian distribution with $\mu=1$ and $\sigma=0.2$~\cite{Um_2017}. 
    In this way, the time series are scaled by a single multiplier for all time steps. 
    \item \textbf{Magnitude Warping}: For the magnitude warping evaluation, we use the augmentation method proposed by~\cite{Um_2017}. In this method, the magnitudes of the time series are multiplied by a warping amount determined by a cubic spline line with four knots at random locations and magnitudes. The knots have peaks or valleys with $\mu=1$ and $\sigma=0.2$.
    \item \textbf{Permutation}: For this implementation, we use permutation with two to five equal sized segments~\cite{Um_2017}. 
    \item \textbf{Slicing}: Slicing, specifically window slicing~\cite{le2016data}, crops the time series to 90\% of the original length. The starting point of the window slice is chosen at random. Also, in order to be directly compared to the other data augmentation methods, we interpolate the time series back to the original length.
    \item \textbf{Time Warping}: 
    The warping path is defined by a smooth cubic spline-based curve with four knots. 
    The knots have random magnitudes with $\mu=1$ and a $\sigma=0.2$~\cite{Um_2017}.
    \item \textbf{Window Warping}: 
    As outlined in~\cite{le2016data}, our Window Warping implementation selects a random window, that is 10\% of the original time series length and warps the time dimension by 0.5 times or 2 times.
    \item \textbf{SuboPtimAl Warped time series geNEratoR (SPAWNER)}:
    SPAWNER~\cite{Kamycki_2019} is a pattern mixing data augmentation method that ``suboptimally'' averages two intra-class randomly selected patterns. 
    We use the standard symmetric slope constraint for DTW with a warping path boundary constraint of 10\% of the time series length. In addition, SPAWNER adds noise with $\mu=0$ and $\sigma=0.5$ in order to further transform the data. 
    \item \textbf{Weighted DTW Barycentric Averaging (wDBA)}: 
    wDBA exploits traditional DBA for the use of data augmentation by weighting patterns in the average. 
    We use the Average Selected with Distance (ASD) version due to it showing the best results~\cite{Forestier_2017}. A symmetric slope constraint is used as DTW's slope constraint and 10\% of the length is used as the warping window.
    \item \textbf{Random Guided Warping (RGW)}: 
    RGW selects random intra-class patterns and warps the elements of one pattern to the time steps of the other. 
    As in~\cite{iwana2020time}, we use the RGW-D version which uses standard DTW with a symmetric slope constraint and a warping path boundary constraint of 10\% of the time series length.
    \item \textbf{Discriminative Guided Warping (DGW)}: 
    DGW is similar to RGW, except it selects the most discriminative pattern in a batch as the teacher~\cite{iwana2020time}. 
    For this implementation, DGW-sD is used. DGW-sD extends the standard DGW-D but uses shapeDTW instead of the typical DTW. This is used for the evaluation as the results were the most competitive variation of guided warping~\cite{iwana2020time}.
\end{itemize}

\section{Results and discussion}
\label{sec:discussion}

The average results of each data augmentation method and each model on the 128 datasets in the 2018 UCR Time Series Archive are shown in Tables~\ref{tab:results_mag}, \ref{tab:results_time}, and~\ref{tab:results_pm}.
In addition to the average results for each data augmentation method and model combination, the tables include a paired $t$-test which compares the differences between no augmentation and each augmentation method. 
Specifically, the $t$-value is shown along with indicators for low two-tailed $p$-values.

\begin{table}[!ht]
\begin{adjustwidth}{-1.5in}{0in} 
\centering
\caption{
{\bf Comparative results for magnitude domain transformation-based data augmentation methods}}
\label{tab:results_mag}
\begin{tabular}{|l+l|l|l|l|l|l|l|l|l|}
\hline
& {\bf None} & \multicolumn{2}{l|}{{\bf Jittering}} &  \multicolumn{2}{l|}{{\bf Rotation}} &  \multicolumn{2}{l|}{{\bf Scaling}} & \multicolumn{2}{l|}{{\bf Magnitude Warping}} \\ 
{\bf Model} & Ave. (\%) & Ave. (\%) & $t$ & Ave. (\%) & $t$ & Ave. (\%) & $t$ & Ave. (\%) & $t$  \\ \thickhline
MLP & 70.23$\pm$21.91 & 70.52$\pm$21.67 & 1.55 & 69.13$\pm$21.70 & -1.95* & 70.24$\pm$21.97 & 0.05 & 69.43$\pm$22.50 & -2.91*** \\ \hline
VGG & 73.02$\pm$23.05 & 74.21$\pm$22.57 & 0.91 & 71.87$\pm$22.14 & -0.84 & 73.69$\pm$23.86 & 0.53 & 74.42$\pm$22.63 & 1.45 \\ \hline
ResNet& 81.39$\pm$17.19 & 80.87$\pm$18.08 & -0.87 & 78.01$\pm$19.93 & -3.31*** & 81.92$\pm$16.81 & 1.05 & 81.33$\pm$17.15 & -0.08 \\ \hline
LSTM& 53.46$\pm$28.10 & 54.72$\pm$27.10 & 1.09 & 49.64$\pm$26.93 & -2.95*** & 53.69$\pm$26.94 & 0.21 & 54.35$\pm$27.29 & 0.93 \\ \hline
BLSTM & 62.51$\pm$24.74 & 60.23$\pm$24.87 & -2.01** & 57.76$\pm$24.76 & -3.51*** & 62.13$\pm$25.89 & -0.37 & 62.00$\pm$26.51 & -0.44 \\ \hline
LSTM-FCN & 81.54$\pm$17.53 & 79.18$\pm$19.87 & -4.89*** & 78.87$\pm$19.37 & -3.66*** & 80.73$\pm$18.37 & -1.74* & 79.46$\pm$19.68 & -3.15*** \\ \hline
\end{tabular}
\begin{flushright} * $p<0.1$, ** $p<0.05$, *** $p<0.01$\end{flushright}
\end{adjustwidth}
\end{table}

\begin{table}[!ht]
\begin{adjustwidth}{-1.51in}{0in} 
\centering
\caption{
{\bf Comparative results for time domain transformation-based data augmentation methods}}
\label{tab:results_time}
\begin{tabular}{|l+l|l|l|l|l|l|l|l|l|}
\hline
 & \multicolumn{2}{l|}{{\bf Permutation}} & \multicolumn{2}{l|}{{\bf Slicing}} & \multicolumn{2}{l|}{{\bf Time Warping}} & \multicolumn{2}{l|}{{\bf Window Warping}} \\ 
{\bf Model} &  Ave. (\%) & $t$ & Ave. (\%) & $t$ & Ave. (\%) & $t$ & Ave. (\%) & $t$  \\  \thickhline
MLP & 68.17$\pm$22.17 & -3.89*** & 70.00$\pm$22.51 & -0.49 & 67.20$\pm$22.31 & -5.64*** & 69.73$\pm$22.50& -0.87 \\ \hline
VGG & 73.75$\pm$22.00 & 0.81 & 76.33$\pm$24.47 & 2.68** & 75.50$\pm$20.82 & 1.86* & 75.88$\pm$21.63 & 2.57**\\ \hline
ResNet&  80.67$\pm$18.31 & -1.16 & 81.51$\pm$17.64 & 0.24 & 79.62$\pm$19.20 & -2.25** & 82.32$\pm$16.93 & 1.99**\\ \hline
LSTM& 49.65$\pm$26.90 & -3.56*** & 52.44$\pm$27.81 & -0.96 & 51.09$\pm$27.14 & -2.28** & 54.41$\pm$28.23 & 1.73*\\ \hline
BLSTM & 60.89$\pm$24.73 & -1.38 & 64.60$\pm$24.01 & 1.93* & 61.42$\pm$25.22 & -0.97 & 62.63$\pm$24.40 & 0.10\\ \hline
LSTM-FCN & 81.25$\pm$17.22 & -0.74 & 80.30$\pm$19.27 & -1.67* & 78.82$\pm$18.69 & -4.66*** & 79.73$\pm$19.35 & -1.94* \\ \hline
\end{tabular}
\begin{flushright} * $p<0.1$, ** $p<0.05$, *** $p<0.01$\end{flushright}
\end{adjustwidth}
\end{table}

\begin{table}[!ht]
\begin{adjustwidth}{-1.63in}{0in} 
\centering
\caption{
{\bf Comparative results for pattern mixing-based data augmentation methods}}
\label{tab:results_pm}
\begin{tabular}{|l+l|l|l|l|l|l|l|l|l|}
\hline
&  \multicolumn{2}{l|}{{\bf SPAWNER}} & \multicolumn{2}{l|}{{\bf wDBA}} & \multicolumn{2}{l|}{{\bf RGW}} & \multicolumn{2}{l|}{{\bf DGW}} \\ 
{\bf Model} & Ave. (\%) & $t$ & Ave. (\%) & $t$ & Ave. (\%) & $t$ & Ave. (\%) & $t$  \\ \thickhline
MLP &  69.15$\pm$22.07 & -2.84*** & 69.49$\pm$21.68 & -2.15** & 70.01$\pm$22.09 & -0.57 & 69.60$\pm$22.70 & -1.19\\ \hline
VGG &  75.77$\pm$21.18 & 2.28** & 73.49$\pm$22.33 & 0.36 & 74.51$\pm$22.18 & 1.54 & 75.78$\pm$22.20 & 2.25**\\ \hline
ResNet&  80.47$\pm$17.02 & -1.25 & 81.46$\pm$18.59 & 0.12 & 81.25$\pm$18.10 & -0.22 & 81.99$\pm$17.38 & 1.29 \\ \hline
LSTM&  54.97$\pm$28.33 & 1.73* & 52.21$\pm$27.29 & -1.19 & 54.04$\pm$28.17 & 0.51 & 54.48$\pm$28.52 & 1.06 \\ \hline
BLSTM &  60.53$\pm$25.66 & -1.50 & 61.41$\pm$24.81 & -1.12 & 62.41$\pm$26.09 & -0.10 & 64.82$\pm$24.34 & 2.21** \\ \hline
LSTM-FCN  & 79.06$\pm$18.57 & -4.75*** & 79.90$\pm$18.93 & -3.70*** & 78.55$\pm$20.16 & -3.09*** & 80.25$\pm$19.94 & -1.49 \\ \hline
\end{tabular}
\begin{flushright} * $p<0.1$, ** $p<0.05$, *** $p<0.01$\end{flushright}
\end{adjustwidth}
\end{table}

The tables show that the different data augmentation methods have dramatic differences in results depending on the neural network architecture. 
Overall, there were mixed results. Some data augmentation methods improved the accuracy and some methods were detrimental. 
Slicing, Window Warping, and DGW tended to have the most positive effects for each model while Rotation, Permutation, and Time Warping had significantly degraded accuracies.
Furthermore, every data augmentation tended to improve the VGG model the most with the largest gain from using Slicing with VGG.
It is also notable that the data augmentation with MLP and LSTM-FCN was mostly detrimental, sometimes significantly. 

\subsection{Differences between augmentation methods}

\begin{figure}[!h]
\centering
\includegraphics{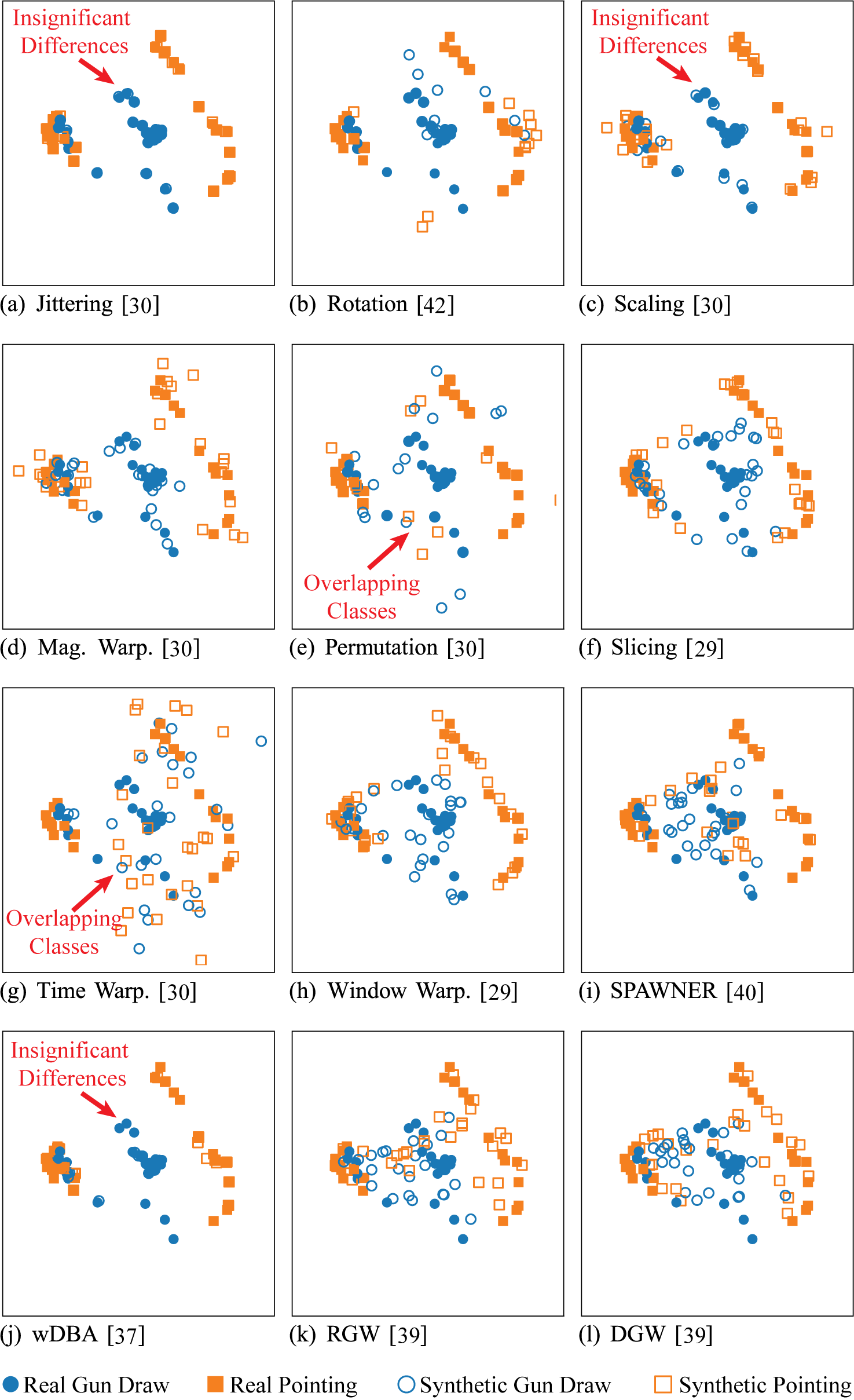}
\caption{{\bf Visualization of the GunPoint dataset using PCA and the compared augmentation methods.} The solid shapes are the original time series, the hollow shapes are the generated time series, and the colors indicate different classes. }
\label{fig:mds}
\end{figure}

Some of the accuracy differences can be explained by examining the effects that the data augmentation has on the data. 
Fig.~\ref{fig:mds} shows a comparison of the data augmentation methods using Principal Component Analysis~(PCA). 
In the figure, the training set of a sample dataset, GunPoint from the 2018 UCR Time Series Archive, is visualized using PCA. 
The solid points are the original training set points and the hollow points are the generated samples. 
The two colors represent the different classes. 
By comparing the generated patterns when projected into the first two principal axes, we can infer some differences between the generated patterns of the data augmentation methods.

For example, wDBA and Jittering created time series that were very similar to the original time series in the GunPoint dataset. 
This is unsurprising since the former weights similar patterns more when mixing and the latter only adds noise. 
The change in accuracy reflected this with Tables~\ref{tab:results_mag} and~\ref{tab:results_pm} demonstrating that wDBA and Jittering only had minor effects on the accuracy (with exception to LSTM-FCN which had significant losses in accuracy for all data augmentation methods). 
Furthermore, the data augmentation methods which transform the patterns so much that the classes are overlapped in Fig.~\ref{fig:mds}, such as Time Warping, Permutation, and Rotation, reflected significant losses in accuracy.
Conversely, as expected, the data augmentation methods which push the boundaries of the classes in the PCA space, tended to perform better.


\subsection{Relationship between dataset properties and accuracy}

In order to understand the differences between the data augmentation methods, we analyze the relationships between the methods and different properties, or characteristics, of the datasets. 
To do this, we find the correlation between the dataset properties and the change in accuracy $\Delta Acc$ from the un-augmented model to the augmented models. 
The following dataset properties are used:
\begin{itemize}
    \item \textbf{Training set size}: The number of training patterns in each dataset.
    \item \textbf{Patterns per class}: The average number of training patterns in each class.
    \item \textbf{Time series length}: The maximum number of time steps of the patterns in the training set for each dataset.
    \item \textbf{Dataset variance}: The average variance of the elements of the time series across each dataset.
    \item \textbf{Intra-class variance}: The average variance within each class.
    \item \textbf{Class Imbalance}: The difference between the size of the classes in the training set.
\end{itemize}
Each dataset property and the expectation of each correlation is detailed below. 
These dataset properties are selected because they provide fundamental differences between time series data.

\begin{figure}[!h]
\centering
\includegraphics[width=1\columnwidth]{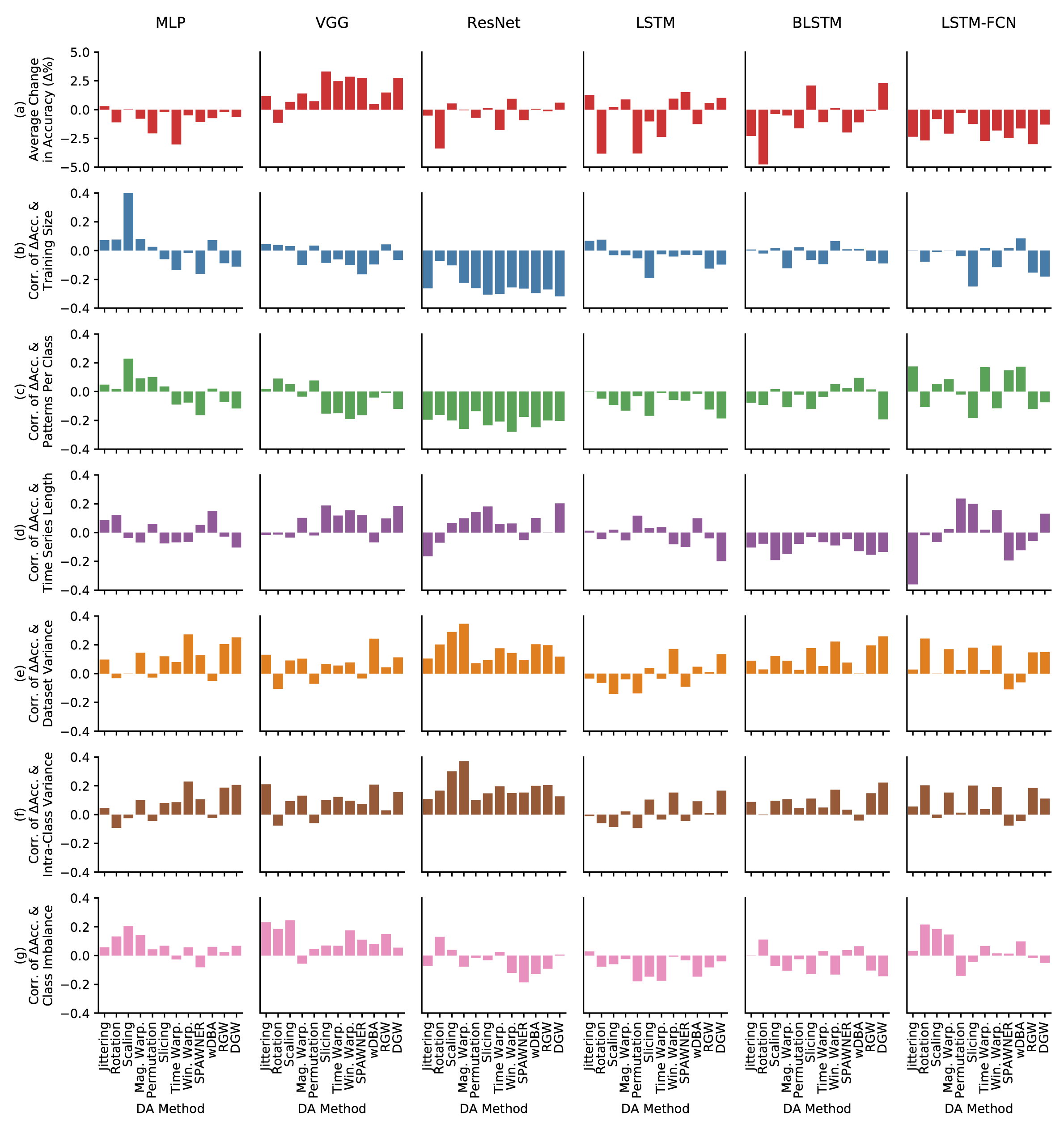}
\caption{{\bf Spearman's Rank Correlation Coefficients between change in accuracy ($\Delta$ Acc) and various dataset characteristics.} The top row in red is the change in accuracy and the subsequent rows are the correlations.}
\label{fig:correlations}
\end{figure}

The results of the correlation analysis are shown in Fig.~\ref{fig:correlations}. 
The first row, Fig.~\ref{fig:correlations}~(a), displays the change in accuracy $\Delta Acc$ for each model and data augmentation method. 
The subsequent rows are the Spearman's Rank Correlation Coefficients between $\Delta Acc$ and each property. 
Spearman's Rank Correlation Coefficient is used because it is robust to outliers and can be used with skewed variable distributions~\cite{mukaka2012guide}. 
As many of the dataset properties is skewed, e.g. most of the datasets have small training sets and most of the datasets have balanced class membership.
However, for reference, Pearson's Product Moment Correlation Coefficient was also used and can be found in \nameref{S1_Fig}.

\subsubsection{Training set size}
The correlation between the change in accuracy $\Delta Acc$ and the number of patterns in the training set is shown in Fig.~\ref{fig:correlations}~(b).
We expect the correlation to be negative because larger datasets are known to already generalize well~\cite{Banko_2001,Torralba_2008}. 
In other words, data augmentation should have a larger effect on smaller datasets, so we would expect a larger increase in accuracy due to data augmentation as the dataset size decreases.

For most model and data augmentation pairs, our expectation is confirmed. 
ResNet and LSTM-FCN especially have strong negative correlations for most of the methods. 
But, even MLP, VGG, and BLSTM had slightly negative correlations between $\Delta Acc$ and training set size.
This implies that for those methods, our expectation is confirmed and the accuracy is inversely related to the number of patterns in the training set. 
However, for some methods, such as Scaling with MLP and LSTM, there is a positive correlation and for many others, there is almost no correlation. 
Thus, for many models, such as MLP, LSTM, and BLSTM, training set size is not a strong indicator for improvement in accuracy. 

\subsubsection{Patterns per class}

The correlation of the change in accuracy to the average number of patterns per class is similar to training set size. 
This property is shown because it is possible for some datasets to have a large number of classes, thus the total training set size is not accurately portrayed. 
Like the total training set size, it would be expected that there is a negative correlation between $\Delta Acc$ and patterns per class. 
However, the results do not differ significantly from the correlation to training set size. 
The correlations between $\Delta Acc$ and patterns per class for MLP, VGG, ResNet, and LSTM were very similar to the correlation between $\Delta Acc$ and training set size. 
Furthermore, in general, the correlations are weaker than the total training set size. 
Thus, consideration based on the other properties or should be used to select the data augmentation method. 

The only major difference from Training Size and Patterns Per Class is LSTM-FCN. 
LSTM-FCN had unexpected positive correlations with Jittering, Time Warping, SPAWNER, and wDBA. 
This indicates that LSTM-FCN might not react as positively to data augmentation when the number of patterns per class is smaller.

\subsubsection{Time series length}

This trial examines the relationship between time series length and change in accuracy. 
The expectation here would be that the correlation between the two is zero because naively, there should be no strong relationship. 
However, from the results in Fig.~\ref{fig:correlations}~(d), it can be observed that there are strong correlations. 
For example, for Slicing, there are positive correlations for the models with CNN components: VGG, ResNet, and LSTM-FCN. 
This means that as the time series grows larger, the gain in accuracy goes up. 
One explanation for this might be due to longer time series containing less information at the endpoints, which Slicing crops. 
Conversely, for the RNN-based networks, LSTM and BLSTM, the correlations were generally negative and LSTM-FCN had a strong negative correlation for Jittering, SPAWNER, and wDBA. 

\subsubsection{Dataset variance}

Next, the correlation between $\Delta Acc$ and the dataset variance $\bar{\sigma}^2_{\mathrm{DS}}$ is found for each data augmentation method and network model combination. 
The dataset variance is the average element-wise variance across the entire training dataset, or:
\begin{equation}
    \bar{\sigma}^2_{\mathrm{DS}}= \frac{1}{T}\sum^{T}_{t=1}\sigma^2_{t},
\end{equation}
where $T$ is the number of time steps and $\sigma^2_t$ is the variance at time step $t$ across the whole dataset. 
The correlation between $\Delta Acc$ and $\bar{\sigma}^2_{\mathrm{DS}}$ explains the relationship between the change and accuracy and the differences between the patterns in the dataset. 
A positive correlation is expected because datasets with small variations have similar patterns, even between classes, and would be disturbed from the transformations of the data augmentation methods. 

From Fig.~\ref{fig:correlations}~(e), we find that the expected positive correlations were observed for all of the models and most of the data augmentation, especially the CNN-based models. 
The only outlier is LSTM which had negative correlations for magnitude domain augmentation methods.

\subsubsection{Intra-class variance}

Next, we use the intra-class variance, or the variance within classes, as a property of the dataset. 
The intra-class variance $\bar{\sigma}^2_{IC}$ is the average element-wise variance, or:
\begin{equation}
    \bar{\sigma}^2_{IC}= \frac{1}{NC}\sum^{C}_{c=1}\frac{N_c}{T}\sum^{T}_{t=1}\sigma^2_t,
\end{equation}
where $C$ is the number of classes, $N$ is the total number of training patterns, $N_c$ is the number of patterns in class $c$, $T$ is the number time steps, and $\sigma^2_{i}$ is the variance of the patterns in class $c$ at time step $i$.
Similar to the dataset variance, the expected correlation between $\bar{\sigma}^2_{IC}$ and $\Delta Acc$ is positive.

The correlations are shown in Fig.~\ref{fig:correlations}~(f) and it shows that the correlations between intra-class variance and change in accuracy act similar to the overall dataset variance. 
ResNet especially has a strong correlation between intra-class variation and change in accuracy. 

\subsubsection{Class imbalance}

Finally, we consider the class imbalance as a property. 
Class imbalance is how imbalanced the size of the classes in the training set is. 
Augmentation methods such as SMOTE~\cite{Chawla_2002} attempt to fix class imbalance by generating extra patterns in the minority classes.
To measure class imbalance, Imbalance-Degree~(ID) proposed by Ortigosa Hernandez~\cite{Ortigosa_Hern_ndez_2017}
is used. 
ID is a robust class imbalance measure that is designed for multi-class problems, much like the datasets used in the experiments. 
ID is calculated 
\begin{equation}
    \label{eq:id}
    ID=\frac{d(\boldsymbol{\zeta},\boldsymbol{\beta})}{d(\boldsymbol{\iota},\boldsymbol{\beta})}+(C_m-1),
\end{equation}
where $d(\cdot,\boldsymbol{\beta})$ is a statistical distance between distributions $\boldsymbol{\zeta}$ or $\boldsymbol{\iota}$ and $\boldsymbol{\beta}$. 
$\boldsymbol{\zeta}=[\zeta_1,\dots,\zeta_c,\dots,\zeta_C]$ denotes a vector of the true distributions $\zeta_c$ of the classes $c$ in the dataset where:
\begin{equation}
    \zeta_c=\frac{N_c}{N},
\end{equation}
and $N_c$ is the number of patterns in each class $c$ and $N$ is the total number of training patterns.
Vector $\boldsymbol{\beta}=[\beta_1,\dots,\beta_c,\dots,\beta_C]$ is the distribution of a balanced dataset with each $\beta_c=\frac{1}{C}$.
$C_m$ is the number of minority classes. A minority class is a class that has with $\zeta_c<\frac{1}{C}$.
Finally, $\boldsymbol{\iota}=[\iota_1,\dots,\iota_c,\dots,\iota_C]$ is a vector representing the class distributions that is has the maximal distance from $\boldsymbol{\beta}$ with the same $C_m$, or a vector of $C_m$ number of $\iota_c=0$, $C - C_m - 1$ number of $\iota_c=\frac{1}{C}$, and one $\iota_c=1-\frac{C-C_m-1}{C}$.
A balanced dataset would have an $ID=0$ and unbalanced datasets have larger ID scores. 
In the experiment, the Hellinger distance is used as the distance measure $d(\cdot,\boldsymbol{\beta})$ due to it having the highest Pearson correlation coefficient between imbalance and neural network performance among the distance functions tested in Ortigosa-Hernandez et al.~\cite{Ortigosa_Hern_ndez_2017}

Unlike SMOTE, the data augmentation methods in the experiments sample the classes at the same distribution as what already exists in the training dataset. 
While small classes can benefit from more training samples, the large classes also increase in size. 
Despite this, there are correlations that can be found from the results.

In Fig.~\ref{fig:correlations}~(g), there are positive correlations for Jittering, Rotation, Scaling, and Magnitude Warping for MLP and LSTM-FCN, and Jittering, Rotation, and Scaling for VGG. 
This is interesting to note because these methods are magnitude domain augmentations. 
In other words, this shows that as class imbalance rises, the $\Delta Acc$ tends to rise for magnitude domain data augmentation methods for MLP, VGG, and LSTM-FCN.
Conversely, the time domain data augmentations tend have negative or very small correlations for the same models.

\subsection{Computation time}

The theoretical computational complexity of many of the methods are similar; the simple transformations are $O(T)$ and the complex transformations and pattern mixing methods are $O(T^2)$, where $T$ is the number of time steps. 
However, as shown in Table~\ref{tab:properties}, the observed execution time varies a lot between the methods. 
To compare the execution time, all of the datasets in the 2018 UCR Time Series Archive were augmented once and timed using a computer with a 2.60 GHz Intel Xeon CPU using the Python implementation at \url{https://github.com/uchidalab/time_series_augmentation}.

\begin{table}[!ht]
\centering
\caption{
{\bf Algorithm Comparison With Average Augmentation Time Per Dataset and Tunable Parameters}}
\label{tab:properties}
\begin{tabular}{|l+l|l|}
\hline
{\bf DA Method} & {\bf Average Time (s)} & {\bf Tunable Parameters} \\  \thickhline
Jittering & $<0.01$ & $\sigma$  \\ \hline 
Rotation & $<0.01$ & $\sigma$ \\ \hline 
Scaling & $<0.01$ & $\sigma$ \\ \hline 
Magnitude Warping & $0.11$ & $\sigma$, \# of knots \\ \hline 
Permutation & 0.01 & window size, \# of permutations \\ \hline 
Slicing & 0.02 & window size \\ \hline 
Time Warping & $0.12$ & $\sigma$, \# of knots\\ \hline 
Window Warping & $0.04$ & window size, warping amount  \\ \hline 
SPAWNER & $66.7$ & $\sigma$, DTW constraints \\ \hline 
wDBA &  2,300 & weighting, DTW constraints \\ \hline 
RGW & 70.5 & DTW constraints \\ \hline 
DGW & 6,380 & batch size, DTW constraints \\ \hline 
\end{tabular}
\end{table}

The average observed execution times for the transformation-based methods are negligible. 
They only took a fraction of a second on average to double the dataset size. 
On the other hand, the pattern mixing methods are much slower with SPAWNER and RGW taking about a minute on average to execute and wDBA and DGW taking 2,300 and 4,290 seconds, respectively. 
The reason for the slow run times is primarily due to the speed of DTW, which each of the methods relies on for element alignment. 
For most datasets, this is not a problem, but ones with long time series, such as HandOutlines with 2,709 time steps, can take extraordinary amounts of time compared to the transformation-based methods.
Accordingly, pattern mixing methods which do not use DTW, such as interpolation, might not face the same issue.
DGW and wDBA take extra long because for each generated time series, multiple DTW calculations must be performed.
In addition, the DGW implementation used in the experiment uses shapeDTW which takes significantly longer to execute than standard DTW.

\subsection{Number of tunable parameters}
The number of tunable parameters is another aspect of the data augmentation methods that one needs to consider. 
The hyperparameter, design choice, and variation of the method need to be selected and the effectiveness of the augmentation method can depend on the parameters. 
Thus, methods with many parameters may need many adjustments and evaluations to be effectively used. 
A list of the parameters that need to be defined manually is displayed in Table~\ref{tab:properties}.
In general, while the random transformations have fewer parameters, they are more dependent on the hyperparameters due to the random element. 
While more complex, the pattern mixing methods on the other hand rely on the patterns in the dataset for randomness, thus, they have fewer choices to tune.

\subsection{Recommendations on data augmentation usage}

Each data augmentation method has different effects depending on the model and dataset, as shown in Tables~\ref{tab:results_mag}, \ref{tab:results_time}, \ref{tab:results_pm}, and Fig.~\ref{fig:correlations}.
Thus, it is an arduous task to determine which data augmentation method to use in which situation. 
In this section, we will attempt to provide recommendations to solve this problem.

\subsubsection{Dataset type and data augmentation method}

The datasets can be broken into categories, such as ECG, sensor, etc. 
To categorize the datasets, we use the labels provided by~\cite{UCRArchive2018}. 
In Table~\ref{tab:recommendation}, we show the methods with the top 5 highest average rank for each category and model combination. 
In the table, execution time is the deciding factor for instances of ties. 
This table can be used as a general guide in combination with Fig.~\ref{fig:correlations}.

\begin{table}[!ht]
\begin{adjustwidth}{-1.5in}{0in} 
\centering
\caption{{\bf Data augmentation recommendations for data type and model type.} }
\label{tab:recommendation}
\footnotesize
\begin{tabular}{|l+l|l|l|l|l|l|}
\hline
Data Type & MLP & VGG & ResNet & LSTM & BLSTM & LSTM-FCN \\ \thickhline
Device & Sl, Ro, Sc, J, N & Sl, M, W, T, RW & N, W, T, P, SP & RW, J, W, Sc, T & J, T, RW, DW, wD & T, P, M, N, RW \\ \hline
ECG & M, Ro, Sc, J, N, SP & Sc, Ro, M, W, RW & J, Sc, N, P, M & Sc, J, Ro, Sl, W & Sl, Sc, N, Ro, J & Sc, M, P, N, wD \\ \hline
EOG & W, wD, RW, J, DW & SP, W, P, T, wD & DW, P, RW, T, SP & Sc, Sl, W, J, P & J, SP, N, T, M & DW, Sc, W, Ro, Sl \\ \hline
EPG & N, Sc, J, Sl, M & J, SP, Sc, T, DW & N, Sc, J, Sl, P & Sc, J, Sl, W, P & N, Sc, J, Sl, P & N, Sc, J, Sl, W \\ \hline
Hemo & RW, Ro, M, N, P & T, P, DW, Sl, W & Sl, Ro, SP, wD, N & T, Sl, P, DW, N & Sl, DW, P, SP, Ro & Sl, N, P, DW, T \\ \hline
HRM & Sl, DW, N, P, RW & T, Sl, Ro, W, P & Sl, P, W, RW, SP & DW, J, SP, Ro, Sl & Sl, RW, P, SP, Ro & N, P, W, RW, wD \\ \hline
Image & Sc, J, Ro, RW, N & W, DW, RW, Sl, T & W, N, Sc, M, wD  & RW, M, SP, W, J & DW, Sc, M, T, W & N, W, P, Sl, DW \\ \hline
Motion & W, DW, Sl, RW, Sc & DW, W, RW, Sl, T & Sc, wD, Sl, W, Ro& W, N, M, RW, J & RW, Sl, SP, W, RW & Sc, W, N, DW, M \\ \hline
Power & N, Sc, Ro, wD, J & N, Sc, Ro, J, M & RW, Ro, SP, wD, W & N, Ro, RW, M, SP & Ro, DW, N, J, RW & Sc, J, Ro, wD, N \\ \hline
Simulated & DW, SP, Sc, RW, J & W, DW, Sc, Sl, N & Sc, DW, W, N, wD & W, DW, Sl, Sc, J & RW, DW, W, M, wD & W, DW, N, Ro, RW \\ \hline
Spectro & Sl, W, DW, RW, T & Sl, W, DW, wD, J & W, DW, RW, SP, Sl & DW, SP, M, T, J & M, T, DW, Sc, SP & W, RW, Sl, Sc, N \\ \hline
Spectrum & J, Sc, RW, N, W & Sl, N, DW, Ro, P & W, DW, Sl, SP, M & Sc, SP, T, RW, W & W, T, J, SP, N & P, SP, Sc, RW, N \\ \hline
Traffic & Sl, Sc, RW, DW, wD & Sc, J, Ro, T, wD & P, wD, J, Ro, N & W, T, Sc, T, RW & W, wD, Sl, DW, N & N, W, J, SP, wD \\ \hline
Trajectory & J, Sc, M, DW, SP & RW, N, DW, Sc, W & Sc, M, wD, W, RW & M, Sc, J, Sl, DW & DW, W, J, Ro, Sc & P, RW, Sc, N, DW \\ \hline
\hline
Overall & Sc, J, RW, W, Sl & W, DW, Sl, RW, T & W, Sc, DW, N, wD & W, RW, J, Sc, M & DW, Sl, RW, W, M & N, W, DW, P, Sc \\
\hline
\end{tabular}
\begin{flushleft} 
The suggested augmentation method has the top five highest average rank within each data type. Ties between methods are broken based on execution time.

N: None, J: Jittering, Sc: Scaling, Ro: Rotation, M: Magnitude Warping, P: Permutation, T: Time Warping, Sl: Slicing, W: Window Warping, SP: SPAWNER, wD: wDBA, RW: RGW, DW: DGW
\end{flushleft}
\end{adjustwidth}
\end{table}

\subsubsection{Magnitude domain transformations}

Jittering, Magnitude Warping, and Scaling had similar results. 
They tend to act similarly since they are similar transformations in that they only differ in how many directions magnitude gets scaled. 
For example, in general, these magnitude domain transformations work well with VGG and with LSTM. 

As described previously, Rotation (flipping in this implementation), seems to be not suitable as a general time series data augmentation method. 
The overall accuracy for Rotation, in Table~\ref{tab:results_mag}, shows a decrease in accuracy for all models. 
Furthermore, in Table~\ref{tab:recommendation}, Rotation only had the highest average rank for one combination, power data with BLSTM. 
This was the second worst showing next to Permutation in Table~\ref{tab:recommendation}. 
The poor performance of Rotation is intuitive, however. 
The 2018 UCR Time Series Archive contains many shape-based datasets and datasets where flipping is inappropriate. 

\subsubsection{Time domain transformations}

Similar to Rotation, Permutation had severely detrimental effects on accuracy. 
However, this is somewhat expected because Permutation breaks the time dependency of the time series.
The only time that it would intuitively make sense for Permutation to be used would be for periodic time series or very sparse time series.

As a general purpose data augmentation method for time series, the other time domain transformations far outperform Permutation. 
Specifically, Slicing and Window Warping performed well on most datasets and models. 
In particular, the CNN-based models, VGG and ResNet, were significantly improved by Slicing and Window Warping (Table~\ref{tab:results_time}). 
Considering the positive effect they have as data augmentation methods and the very fast computation time (Table~\ref{tab:properties}), it seems like they should be a first choice when selecting time series data augmentation.

However, Time Warping showed a poor performance. 
This is likely due to over transforming the time series causing significant noise, as illustrated in Fig.~\ref{fig:mds}~(g).
The parameters used for this implementation of Time Warping were the parameters suggested by Um et al.~\cite{Um_2017}. 
This reinforces the flaw with random transformation-based data augmentation methods in which the parameters must be carefully selected. 
The difficulty is that the user has to balance transforming the patterns enough so that the generalization is increased, but not so much that the classes are confused. 

\subsubsection{Pattern matching methods}

As mentioned previously, the largest downside to the pattern matching methods is the slow computation time. 
While it is possible to achieve better generalization and results using these methods, one has to consider the extra time it would take to generate the data. 
However, this issue is only with longer patterns and the execution time is also negligible for short time series. 
Furthermore, from Fig.~\ref{fig:correlations}~(d), the correlation between change in accuracy and time series length is negative for these methods. 
Thus, using these pattern mixing methods would be recommended for shorter time series.

However, for most datasets, wDBA had disappointing results. 
The primary reason for this is due to wDBA not creating diverse enough results. 
In the evaluation, we used the ASD because Forestier et al.~\cite{Forestier_2017} found it the most competitive weighting method. 
The ASD weights the nearest neighbors of the reference pattern more than farther away patterns. 
Due to this, as shown in Fig.~\ref{fig:mds}~(j), the new patterns generated from wDBA were not significantly different from the existing patterns in the dataset. 
Although, it should be noted that this could be an issue with the ASD weighting scheme. 

As for SPAWNER, RGW, and DGW, there were mixed results. 
In some models and some datasets, they performed better than other augmentation methods and on others, they performed worse. 
Notably, DGW had the most performance increase compared to no augmentation out of all of the pattern mixing methods and the highest average rank on BLSTM compared to all augmentation methods (Table~\ref{tab:recommendation}). 
As mentioned before, the downside of DGW is that it is also the slowest algorithm of all the data augmentation methods used for the comparative evaluations. 
For datasets with many long time series, such as HandOutlines, StarLightCurves, and NonInvasiveFetalECGThorax1/2, this could mean many orders of magnitude longer than the simple transformations. 

\section{Conclusion}
\label{sec:conc}

In this paper, we performed a comprehensive survey of data augmentation methods for time series. 
The survey categorizes and outlines the various time series data augmentation methods. 
We include transformation-based methods across the related time series domains and time series pattern mixing, generative models, and decomposition methods. 
Furthermore, a taxonomy of time series data augmentation methods was proposed. 

In addition, an empirical comparative evaluation of 12 time series data augmentation methods on six neural network models and 128 discrete finite time series datasets was performed. 
Namely, we use all of the datasets in the 2018 UCR Time Series Archive~\cite{UCRArchive2018} to evaluate jittering, rotation, scaling, magnitude warping, permutation, slicing, time warping, window warping, SPAWNER, wDBA, RGW, and DGW. 
The training datasets of each dataset are augmented by four times the size and are trained and tested using an MLP, VGG, ResNet, LSTM, BLSTM, and LSTM-FCN. 
Through the empirical evaluation, we are able to compare the data augmentations and analyze the findings.

By using all 128 datasets of the 2018 UCR Time Series Archive, we are able to test the data augmentation methods on a wide variety of time series and make recommendations on data augmentation usage. 
As a general, easy-to-use, and effective method, the Window Warping algorithm that was proposed by Le Guennec et al.~\cite{le2016data} is the most recommended algorithm. 
Window warping had the highest average rank across all of the datasets for VGG, ResNet, and LSTM and consistently performed well with the other datasets as well.
In addition, slicing, i.e., window slicing proposed by the same authors, also showed to be very effective in most cases. 
While there were algorithms that periodically performed better than window warping and slicing, the drawbacks of speed or sensitivity to parameter tuning hindered the other methods.
Alternatively, we found that the time domain pattern mixing method, DGW, also performed well in most cases. However, it is only recommended on time series that are shorter in length due to the high computational requirement.

The results also revealed some key aspects of data augmentation with time series based neural networks. 
For example, LSTM-FCN does not respond well to data augmentation. 
Part of this could be because LSTM-FCN uses dropout with a very high rate or it could be that the design of the architecture is just not suitable for data augmentation.
MLP, to a lesser extent, also did not respond well.
Conversely, the accuracy of VGG was often improved with most of the data augmentation methods. 
The other models, ResNet, LSTM, and BLSTM had mixed results, and careful augmentation method selection is required.

We also analyzed different aspects of time series datasets and the effects data augmentation had on datasets with them. 
First, the correlations between properties of time series datasets and the change in accuracy from augmentation were found. 
The findings showed that there generally was a negative correlation between change in accuracy and training set size (and number of patterns per class) and a positive correlation for dataset variance and intra-class variance. 
We found that time series length generally had a negative correlation for RNN-based models, but positive correlations for CNN-based models.
There was a strong positive correlation between change in accuracy and class imbalance for MLP, VGG, and LSTM-FCN for the magnitude domain data augmentation methods. 
Second, using ranking, we found the top augmentation methods for each model and dataset type combination. 

This survey serves as a guide for researchers and developers in selecting the appropriate data augmentation method for their applications.
Using this, it is possible to refine the selection of data augmentation based on dataset type, property, and model.
An easy to use implementation of the algorithms evaluated in this survey are provided at \url{https://github.com/uchidalab/time_series_augmentation}.

Since data augmentation for time series is not established as much as data augmentation for images, there is a lot of room for time series data augmentation to grow. 
For example, like most works, this survey only uses a single data augmentation method for each model. 
It is possible that multiple data augmentation methods can synergize well and be used serially. With exception to Um et al.~\cite{Um_2017}, this idea is not used. 
There are also many other advanced data augmentation methods used in the image domain~\cite{Shorten_2019} that are not used by time series, such as style transfer, meta-learning, and filters. 
There is a lot of potential for new time series data augmentation research.

\nolinenumbers


\begin{thebibliography}{100}

\bibitem{ding2008querying}
Ding H, Trajcevski G, Scheuermann P, Wang X, Keogh E.
\newblock Querying and mining of time series data.
\newblock Proc Very Larg Data Base Endow. 2008;1(2):1542--1552.
\newblock doi:{10.14778/1454159.1454226}.

\bibitem{Schmidhuber_2015}
Schmidhuber J.
\newblock Deep learning in neural networks: An overview.
\newblock Neural Networks. 2015;61:85--117.
\newblock doi:{10.1016/j.neunet.2014.09.003}.

\bibitem{Wang_2017}
Wang Z, Yan W, Oates T.
\newblock Time series classification from scratch with deep neural networks: A
  strong baseline.
\newblock In: IJCNN; 2017.

\bibitem{rumelhart1988learning}
Rumelhart DE, Hinton GE, Williams RJ.
\newblock Learning representations by back-propagating errors.
\newblock Nat. 1986;323(6088):533--536.
\newblock doi:{10.1038/323533a0}.

\bibitem{Chen_2019comparative}
Chen Q, Liang B, Wang J.
\newblock A comparative study of {LSTM} and phased {LSTM} for gait prediction.
\newblock Int J Artificial Intelli \& App. 2019;10(4):57--66.
\newblock doi:{10.5121/ijaia.2019.10405}.

\bibitem{Kluwak_2020}
Kluwak K, Nizynski T.
\newblock Gait classification using {LSTM} networks for tagging system.
\newblock In: IEEE SoSE; 2020.

\bibitem{Xu_2020}
Xu G, Xing G, Jiang J, Jiang J, Ke Y.
\newblock Arrhythmia detection using gated recurrent unit network with {ECG}
  signals.
\newblock J Medical Imag and Health Inform. 2020;10(3):750--757.
\newblock doi:{10.1166/jmihi.2020.2928}.

\bibitem{Kim_2020}
Kim BH, Pyun JY.
\newblock {ECG} identification for personal authentication using {LSTM}-based
  deep recurrent neural networks.
\newblock Sensors. 2020;20(11):3069.
\newblock doi:{10.3390/s20113069}.

\bibitem{Sun_2016}
Sun L, Su T, Liu C, Wang R.
\newblock Deep {LSTM} networks for online Chinese handwriting Recognition.
\newblock In: ICFHR; 2016.

\bibitem{Carbune_2020}
Carbune V, Gonnet P, Deselaers T, Rowley HA, Daryin A, Calvo M, et~al.
\newblock Fast multi-language {LSTM}-based online handwriting recognition.
\newblock Int J Doc Analy and Recogn. 2020;23(2):89--102.
\newblock doi:{10.1007/s10032-020-00350-4}.

\bibitem{Lecun_1998}
Lecun Y, Bottou L, Bengio Y, Haffner P.
\newblock Gradient-based learning applied to document recognition.
\newblock Proc {IEEE}. 1998;86(11):2278--2324.
\newblock doi:{10.1109/5.726791}.

\bibitem{bai2018empirical}
Bai S, Kolter JZ, Koltun V.
\newblock An empirical evaluation of generic convolutional and recurrent
  networks for sequence modeling.
\newblock arXiv preprint arXiv:180301271. 2018;.

\bibitem{Kenji_Iwana_2020}
Iwana BK, Uchida S.
\newblock Time series classification using local distance-based features in
  multi-modal fusion networks.
\newblock Pattern Recogn. 2020;97:107024.
\newblock doi:{10.1016/j.patcog.2019.107024}.

\bibitem{Banko_2001}
Banko M, Brill E.
\newblock Scaling to very very large corpora for natural language
  disambiguation.
\newblock In: AMACL; 2001.

\bibitem{Torralba_2008}
Torralba A, Fergus R, Freeman WT.
\newblock 80 Million tiny images: A large data set for nonparametric object and
  scene recognition.
\newblock {IEEE} Trans Pattern Anal and Mach Intell. 2008;30(11):1958--1970.
\newblock doi:{10.1109/tpami.2008.128}.

\bibitem{UCRArchive2018}
Dau HA, Keogh E, Kamgar K, Yeh CCM, Zhu Y, Gharghabi S, et~al.. The UCR time
  series classification archive; 2018.

\bibitem{Shorten_2019}
Shorten C, Khoshgoftaar TM.
\newblock A survey on image data augmentation for deep learning.
\newblock J Big Data. 2019;6(1).
\newblock doi:{10.1186/s40537-019-0197-0}.

\bibitem{olson2018modern}
Olson M, Wyner A, Berk R.
\newblock Modern neural networks generalize on small data sets.
\newblock In: NeurIPS; 2018. p. 3619--3628.

\bibitem{Blagus_2013}
Blagus R, Lusa L.
\newblock {SMOTE} for high-dimensional class-imbalanced data.
\newblock {BMC} Bioinformatics. 2013;14(1).
\newblock doi:{10.1186/1471-2105-14-106}.

\bibitem{hasibi2019augmentation}
Hasibi R, Shokri M, Dehghan M.
\newblock Augmentation scheme for dealing with imbalanced network traffic
  classification using deep learning.
\newblock arXiv preprint arXiv:190100204. 2019;.

\bibitem{krizhevsky2009learning}
Krizhevsky A.
\newblock Learning multiple layers of features from tiny images.
\newblock University of Tront Thesis. 2009;.

\bibitem{ILSVRC15}
Russakovsky O, Deng J, Su H, Krause J, Satheesh S, Ma S, et~al.
\newblock ImageNet large scale visual recognition challenge.
\newblock Int J Comput Vis. 2015;115(3):211--252.
\newblock doi:{10.1007/s11263-015-0816-y}.

\bibitem{simonyan2014very}
Simonyan K, Zisserman A.
\newblock Very deep convolutional networks for large-scale image recognition.
\newblock arXiv preprint arXiv:14091556. 2014;.

\bibitem{He_2016}
He K, Zhang X, Ren S, Sun J.
\newblock Deep residual learning for image recognition.
\newblock In: {IEEE} CVPR; 2016.

\bibitem{Huang_2017}
Huang G, Liu Z, Maaten LVD, Weinberger KQ.
\newblock Densely connected convolutional networks.
\newblock In: IEEE CVPR; 2017.

\bibitem{Szegedy_2015}
Szegedy C, Liu W, Jia Y, Sermanet P, Reed S, Anguelov D, et~al.
\newblock Going deeper with convolutions.
\newblock In: IEEE CVPR; 2015.

\bibitem{wen2020time}
Wen Q, Sun L, Song X, Gao J, Wang X, Xu H.
\newblock Time series data augmentation for deep learning: A survey.
\newblock arXiv preprint arXiv:200212478. 2020;.

\bibitem{Fields_2019}
Fields T, Hsieh G, Chenou J.
\newblock Mitigating drift in time series data with noise augmentation.
\newblock In: ICCSCI; 2019.

\bibitem{le2016data}
{Le Guennec} A, Malinowski S, Tavenard R.
\newblock Data augmentation for time series classification using convolutional
  neural networks.
\newblock In: IWAATD; 2016.

\bibitem{Um_2017}
Um TT, Pfister FMJ, Pichler D, Endo S, Lang M, Hirche S, et~al.
\newblock Data augmentation of wearable sensor data for Parkinson's disease
  monitoring using convolutional neural networks.
\newblock In: ACM ICMI; 2017. p. 216--220.

\bibitem{jaitly2013vocal}
Jaitly N, Hinton GE.
\newblock Vocal tract length perturbation (VTLP) improves speech recognition.
\newblock In: ICML WDLASLP; 2013.

\bibitem{Cao_2014}
Cao H, Tan VYF, Pang JZF.
\newblock A parsimonious mixture of Gaussian trees model for oversampling in
  imbalanced and multimodal time-series classification.
\newblock {IEEE} Trans Neural Networks and Learning Sys.
  2014;25(12):2226--2239.
\newblock doi:{10.1109/tnnls.2014.2308321}.

\bibitem{Wendling_2000}
Wendling F, Bellanger JJ, Bartolomei F, Chauvel P.
\newblock Relevance of nonlinear lumped-parameter models in the analysis of
  depth-{EEG} epileptic signals.
\newblock Bio Cybernetics. 2000;83(4):367--378.
\newblock doi:{10.1007/s004220000160}.

\bibitem{goodfellow2014generative}
Goodfellow I, Pouget-Abadie J, Mirza M, Xu B, Warde-Farley D, Ozair S, et~al.
\newblock Generative adversarial nets.
\newblock In: NeurIPS; 2014. p. 2672--2680.

\bibitem{Bergmeir_2016}
Bergmeir C, Hyndman RJ, Ben{\'{\i}}tez JM.
\newblock Bagging exponential smoothing methods using {STL} decomposition and
  Box{\textendash}Cox transformation.
\newblock Int J Forecasting. 2016;32(2):303--312.
\newblock doi:{10.1016/j.ijforecast.2015.07.002}.

\bibitem{Eltoft}
{Eltoft} T.
\newblock Data augmentation using a combination of independent component
  analysis and non-linear time-series prediction.
\newblock In: IJCNN; 2002. p. 448--453.

\bibitem{Forestier_2017}
Forestier G, Petitjean F, Dau HA, Webb GI, Keogh E.
\newblock Generating synthetic time series to augment sparse datasets.
\newblock In: IEEE ICDM; 2017.

\bibitem{Liu_2020}
Liu B, Zhang Z, Cui R.
\newblock Efficient Time Series Augmentation Methods.
\newblock In: {CISP}-{BMEI}; 2020.

\bibitem{iwana2020time}
Iwana BK, Uchida S.
\newblock Time series data augmentation for neural networks by time warping
  with a discriminative teacher.
\newblock In: ICPR; 2021.

\bibitem{Kamycki_2019}
Kamycki K, Kapuscinski T, Oszust M.
\newblock Data augmentation with suboptimal warping for time-series
  classification.
\newblock Sensors. 2019;20(1):98.
\newblock doi:{10.3390/s20010098}.

\bibitem{Huang_2019}
Huang CL.
\newblock Exploring effective data augmentation with {TDNN}-{LSTM} neural
  network embedding for speaker recognition.
\newblock In: {IEEE} ASRUW; 2019. p. 291--295.

\bibitem{Rashid_2019}
Rashid KM, Louis J.
\newblock Time-warping: A time series data augmentation of {IMU} data for
  construction equipment activity identification.
\newblock In: ISARC; 2019.

\bibitem{Bishop_1995}
Bishop CM.
\newblock Training with noise is equivalent to Tikhonov regularization.
\newblock Neural Computation. 1995;7(1):108--116.
\newblock doi:{10.1162/neco.1995.7.1.108}.

\bibitem{An_1996}
An G.
\newblock The effects of adding noise during backpropagation Training on a
  Generalization Performance.
\newblock Neural Computation. 1996;8(3):643--674.
\newblock doi:{10.1162/neco.1996.8.3.643}.

\bibitem{Arslan_2019}
Arslan M, Guzel M, Demirci M, Ozdemir S.
\newblock {SMOTE} and Gaussian noise based sensor data augmentation.
\newblock In: ICCSE; 2019.

\bibitem{Chawla_2002}
Chawla NV, Bowyer KW, Hall LO, Kegelmeyer WP.
\newblock {SMOTE}: Synthetic minority over-sampling technique.
\newblock J Art Intelli Research. 2002;16:321--357.
\newblock doi:{10.1613/jair.953}.

\bibitem{IsmailFawaz2018}
Ismail~Fawaz H, Forestier G, Weber J, Idoumghar L, Muller PA.
\newblock Data augmentation using synthetic data for time series classification
  with deep residual networks.
\newblock In: IWAATD; 2018.

\bibitem{Ohashi_2017}
Ohashi H, Ahmed S, Akiyama T, Sato T, Nguyen P, Nakamura K, et~al.
\newblock Augmenting wearable sensor data with physical constraint for
  DNN-based human-action recognition.
\newblock In: ICML Workshops; 2017.

\bibitem{Delgado_Escano_2019}
Delgado-Escano R, Castro FM, Cozar JR, Marin-Jimenez MJ, Guil N.
\newblock An end-to-end multi-task and fusion {CNN} for inertial-based gait
  recognition.
\newblock {IEEE} Access. 2019;7:1897--1908.
\newblock doi:{10.1109/access.2018.2886899}.

\bibitem{Tran_2020}
Tran L, Choi D.
\newblock Data augmentation for inertial sensor-based gait deep neural network.
\newblock {IEEE} Access. 2020;8:12364--12378.
\newblock doi:{10.1109/access.2020.2966142}.

\bibitem{Pan_2020}
Pan Q, Li X, Fang L.
\newblock Data augmentation for deep learning-based {ECG} analysis.
\newblock In: Feature Eng. and Comput. Intell. in {ECG} Monitor. Springer;
  2020. p. 91--111.

\bibitem{Steven_Eyobu_2018}
Eyobu OS, Han D.
\newblock Feature representation and data augmentation for human activity
  classification based on wearable {IMU} sensor data using a deep {LSTM} neural
  network.
\newblock Sensors. 2018;18(9):2892.
\newblock doi:{10.3390/s18092892}.

\bibitem{Nagano2019}
{Nagano} T, {Fukuda} T, {Suzuki} M, {Kurata} G.
\newblock Data augmentation based on vowel stretch for improving children's
  speech recognition.
\newblock In: {IEEE} ASRUW; 2019. p. 502--508.

\bibitem{Nguyen_2020}
Nguyen TS, Stuker S, Niehues J, Waibel A.
\newblock Improving sequence-to-sequence speech recognition training with
  on-the-fly data augmentation.
\newblock In: IEEE ICASSP; 2020.

\bibitem{Vachhani_2018}
Vachhani B, Bhat C, Kopparapu SK.
\newblock Data augmentation using healthy speech for dysarthric speech
  recognition.
\newblock In: Interspeech; 2018.

\bibitem{Cui_2014}
Cui X, Goel V, Kingsbury B.
\newblock Data augmentation for deep neural network acoustic modeling.
\newblock In: {IEEE} ICASSP; 2014.

\bibitem{Lee_1998}
Lee L, Rose R.
\newblock A frequency warping approach to speaker normalization.
\newblock {IEEE} Trans Speech and Audio Process. 1998;6(1):49--60.
\newblock doi:{10.1109/89.650310}.

\bibitem{Adachi_2007}
Adachi S, Takemoto H, Kitamura T, Mokhtari P, Honda K.
\newblock Vocal tract length perturbation and its application to male-female
  vocal tract shape conversion.
\newblock J Acousitical Soc of America. 2007;121(6):3874--3885.
\newblock doi:{10.1121/1.2730743}.

\bibitem{ko2015audio}
Ko T, Peddinti V, Povey D, Khudanpur S.
\newblock Audio augmentation for speech recognition.
\newblock In: Interspeech; 2015.

\bibitem{Kim_2019}
Kim C, Shin M, Garg A, Gowda D.
\newblock Improved vocal tract length perturbation for a state-of-the-art
  end-to-end speech recognition system.
\newblock In: Interspeech; 2019.

\bibitem{gao2020robusttad}
Gao J, Song X, Wen Q, Wang P, Sun L, Xu H.
\newblock RobustTAD: Robust time series anomaly detection via decomposition and
  convolutional neural networks.
\newblock arXiv preprint arXiv:200209545. 2020;.

\bibitem{Park_2019}
Park DS, Chan W, Zhang Y, Chiu CC, Zoph B, Cubuk ED, et~al.
\newblock {SpecAugment}: A simple data augmentation method for automatic speech
  recognition.
\newblock In: Interspeech; 2019.

\bibitem{Khadijah_2018}
Khadijah, Endah SN, Kusumaningrum R, Rismiyati.
\newblock The study of synthetic minority over-sampling technique ({SMOTE}) and
  weighted extreme learning machine for handling imbalance problem on
  multiclass microarray classification.
\newblock In: ICICoS; 2018.

\bibitem{Bunkhumpornpat_2009}
Bunkhumpornpat C, Sinapiromsaran K, Lursinsap C.
\newblock Safe-level-{SMOTE}: Safe-level-synthetic minority over-sampling
  {technique} for handling the class imbalanced problem.
\newblock In: Adv. in Knowl. Disc. and Data Mining. Springer; 2009. p.
  475--482.

\bibitem{Tarawneh_2020}
Tarawneh AS, Hassanat ABA, Almohammadi K, Chetverikov D, Bellinger C.
\newblock {SMOTEFUNA}: Synthetic minority over-sampling technique based on
  furthest neighbour algorithm.
\newblock {IEEE} Access. 2020;8:59069--59082.
\newblock doi:{10.1109/access.2020.2983003}.

\bibitem{Zhou_2016}
Zhou C, Liu B, Wang S.
\newblock {CMO}-{SMOTE}: Misclassification cost minimization oriented synthetic
  minority oversampling technique for imbalanced learning.
\newblock In: IHMSC; 2016.

\bibitem{Bunkhumpornpat_2011}
Bunkhumpornpat C, Sinapiromsaran K, Lursinsap C.
\newblock {DBSMOTE}: Density-based synthetic minority over-sampling
  {technique}.
\newblock Applied Intell. 2011;36(3):664--684.
\newblock doi:{10.1007/s10489-011-0287-y}.

\bibitem{Gong_2016}
Gong Z, Chen H.
\newblock Model-based oversampling for imbalanced sequence classification.
\newblock In: ACM ICIKM; 2016.

\bibitem{Sawicki_2020}
Sawicki A, Zieli{\'{n}}ski SK.
\newblock Augmentation of segmented motion capture data for improving
  generalization of deep neural networks.
\newblock In: CISIM. Springer; 2020. p. 278--290.

\bibitem{devries2017dataset}
DeVries T, Taylor GW.
\newblock Dataset augmentation in feature space.
\newblock arXiv preprint arXiv:170205538. 2017;.

\bibitem{Yeomans_2019}
Yeomans J, Thwaites S, Robertson WSP, Booth D, Ng B, Thewlis D.
\newblock Simulating time-series data for improved deep neural network
  performance.
\newblock {IEEE} Access. 2019;7:131248--131255.
\newblock doi:{10.1109/access.2019.2940701}.

\bibitem{Savitzky_1964}
Savitzky A, Golay MJE.
\newblock Smoothing and differentiation of data by simplified least squares
  procedures.
\newblock Analytical Chemistry. 1964;36(8):1627--1639.
\newblock doi:{10.1021/ac60214a047}.

\bibitem{sakoe1978dynamic}
Sakoe H, Chiba S.
\newblock Dynamic programming algorithm optimization for spoken word
  recognition.
\newblock IEEE Trans Acoustics, Speech, and Sig Process. 1978;26(1):43--49.

\bibitem{Takahashi_2016}
Takahashi N, Gygli M, Pfister B, Gool LV.
\newblock Deep convolutional neural networks and data augmentation for acoustic
  event recognition.
\newblock In: Interspeech; 2016.

\bibitem{Nanni_2020}
Nanni L, Maguolo G, Paci M.
\newblock Data augmentation approaches for improving animal audio
  classification.
\newblock Ecological Informatics. 2020;57:101084.
\newblock doi:{10.1016/j.ecoinf.2020.101084}.

\bibitem{Petitjean_2011}
Petitjean F, Ketterlin A, Gan{\c{c}}arski P.
\newblock A global averaging method for dynamic time warping, with applications
  to clustering.
\newblock Pattern Recogn. 2011;44(3):678--693.
\newblock doi:{10.1016/j.patcog.2010.09.013}.

\bibitem{smyl2016data}
Smyl S, Kuber K.
\newblock Data preprocessing and augmentation for multiple short time series
  forecasting with recurrent neural networks.
\newblock In: ISF; 2016.

\bibitem{Kang_2020}
Kang Y, Hyndman RJ, Li F.
\newblock {GRATIS}: {GeneRAting} {TIme} Series with diverse and controllable
  characteristics.
\newblock Stat Anal and Data Mining: The {ASA} Data Sci J.
  2020;doi:{10.1002/sam.11461}.

\bibitem{Tanner_1987}
Tanner MA, Wong WH.
\newblock The calculation of posterior distributions by data augmentation.
\newblock J American Stat Assoc. 1987;82(398):528--540.
\newblock doi:{10.1080/01621459.1987.10478458}.

\bibitem{Fr_hwirth_Schnatter_1994}
Fr{\:u}hwirth-Schnatter S.
\newblock Data augmentation and dynamic linear models.
\newblock J Time Series Anal. 1994;15(2):183--202.
\newblock doi:{10.1111/j.1467-9892.1994.tb00184.x}.

\bibitem{Meng_1999}
Meng XL.
\newblock Seeking efficient data augmentation schemes via conditional and
  marginal augmentation.
\newblock Biometrika. 1999;86(2):301--320.
\newblock doi:{10.1093/biomet/86.2.301}.

\bibitem{hou2018sequence}
Hou Y, Liu Y, Che W, Liu T.
\newblock Sequence-to-sequence data augmentation for dialogue language
  understanding.
\newblock arXiv preprint arXiv:180701554. 2018;.

\bibitem{Longpre_2019}
Longpre S, Lu Y, Tu Z, DuBois C.
\newblock An exploration of data augmentation and sampling techniques for
  domain-agnostic question answering.
\newblock In: WMRQA. ACL; 2019.

\bibitem{oord2016wavenet}
Oord Avd, Dieleman S, Zen H, Simonyan K, Vinyals O, Graves A, et~al.
\newblock Wavenet: A generative model for raw audio.
\newblock arXiv preprint arXiv:160903499. 2016;.

\bibitem{Wang_2019wave}
Wang J, Kim S, Lee Y.
\newblock Speech augmentation using Wavenet in speech recognition.
\newblock In: IEEE ICASSP; 2019.

\bibitem{Tu_2018}
Tu J, Liu H, Meng F, Liu M, Ding R.
\newblock Spatial-temporal data augmentation based on {LSTM} autoencoder
  network for skeleton-based human action recognition.
\newblock In: IEEE ICIP; 2018.

\bibitem{Lou_2018}
Lou H, Qi Z, Li J.
\newblock One-dimensional data augmentation using a Wasserstein generative
  adversarial network with supervised signal.
\newblock In: CCDC; 2018.

\bibitem{Haradal_2018}
Haradal S, Hayashi H, Uchida S.
\newblock Biosignal data augmentation based on generative adversarial networks.
\newblock In: EMBC; 2018.

\bibitem{esteban2017real}
Esteban C, Hyland SL, R{\"a}tsch G.
\newblock Real-valued (medical) time series generation with recurrent
  conditional GANs.
\newblock arXiv preprint arXiv:170602633. 2017;.

\bibitem{ramponi2018t}
Ramponi G, Protopapas P, Brambilla M, Janssen R.
\newblock T-CGAN: Conditional generative adversarial network for data
  augmentation in noisy time series with irregular sampling.
\newblock arXiv preprint arXiv:181108295. 2018;.

\bibitem{Che_2017}
Che Z, Cheng Y, Zhai S, Sun Z, Liu Y.
\newblock Boosting deep learning risk prediction with generative adversarial
  networks for electronic health records.
\newblock In: IEEE ICDM; 2017.

\bibitem{Chen_2019}
Chen G, Zhu Y, Hong Z, Yang Z.
\newblock {EmotionalGAN}: Generating ECG to enhance emotion state
  classification.
\newblock In: AICS; 2019.

\bibitem{Hatamian_2020}
Hatamian FN, Ravikumar N, Vesal S, Kemeth FP, Struck M, Maier A.
\newblock The effect of data augmentation on classification of atrial
  fibrillation in short single-lead {ECG} signals using deep neural networks.
\newblock In: {IEEE} ICASSP; 2020.

\bibitem{Madhu_2019}
Madhu A, Kumaraswamy S.
\newblock Data augmentation using generative adversarial network for
  environmental sound classification.
\newblock In: ESPC; 2019.

\bibitem{Zhu_2019}
Zhu F, Ye F, Fu Y, Liu Q, Shen B.
\newblock Electrocardiogram generation with a bidirectional {LSTM}-{CNN}
  generative adversarial network.
\newblock Scientific Reports. 2019;9(1).
\newblock doi:{10.1038/s41598-019-42516-z}.

\bibitem{mirza2014conditional}
Mirza M, Osindero S.
\newblock Conditional generative adversarial nets.
\newblock arXiv preprint arXiv:14111784. 2014;.

\bibitem{nikolaidis2019augmenting}
Nikolaidis K, Kristiansen S, Goebel V, Plagemann T, Liest{\o}l K, Kankanhalli
  M.
\newblock Augmenting physiological time series data: A case study for sleep
  apnea detection.
\newblock In: ECML/PKDD; 2019.

\bibitem{Harada_2019}
Harada S, Hayashi H, Uchida S.
\newblock Biosignal generation and latent variable analysis with recurrent
  generative adversarial networks.
\newblock {IEEE} Access. 2019;7:144292--144302.
\newblock doi:{10.1109/access.2019.2934928}.

\bibitem{Sheng_2019}
Sheng P, Yang Z, Qian Y.
\newblock {GANs} for children: A generative data augmentation strategy for
  children speech recognition.
\newblock In: {IEEE} ASRUW; 2019. p. 129--135.

\bibitem{Wang_2018}
Wang F, hua Zhong S, Peng J, Jiang J, Liu Y.
\newblock Data augmentation for {EEG}-based emotion recognition with deep
  convolutional neural networks.
\newblock In: ICMM; 2018. p. 82--93.

\bibitem{Huang_1998}
Huang NE, Shen Z, Long SR, Wu MC, Shih HH, Zheng Q, et~al.
\newblock The empirical mode decomposition and the Hilbert spectrum for
  nonlinear and non-stationary time series analysis.
\newblock Proc Royal Society of London Series A: Math, Physical and Eng Sci.
  1998;454(1971):903--995.
\newblock doi:{10.1098/rspa.1998.0193}.

\bibitem{Nam_2020}
Nam GH, Bu SJ, Park NM, Seo JY, Jo HC, Jeong WT.
\newblock Data augmentation using empirical mode decomposition on neural
  networks to classify impact noise in vehicle.
\newblock In: IEEE ICASSP; 2020.

\bibitem{Comon_1994}
Comon P.
\newblock Independent component analysis, A new concept?
\newblock Sig Process. 1994;36(3):287--314.
\newblock doi:{10.1016/0165-1684(94)90029-9}.

\bibitem{cleveland1990stl}
Cleveland RB, Cleveland WS, McRae JE, Terpenning I.
\newblock STL: A seasonal-trend decomposition.
\newblock J Official Stat. 1990;6(1):3--73.

\bibitem{Wen_2019}
Wen Q, Gao J, Song X, Sun L, Xu H, Zhu S.
\newblock {RobustSTL}: A robust seasonal-trend decomposition algorithm for long
  time series.
\newblock {AAAI} Conf Artificial Intell. 2019;33:5409--5416.
\newblock doi:{10.1609/aaai.v33i01.33015409}.

\bibitem{He_2015_ICCV}
He K, Zhang X, Ren S, Sun J.
\newblock Delving deep into rectifiers: Surpassing human-level performance on
  ImageNet classification.
\newblock In: IEEE ICCV; 2015.

\bibitem{batchnorm}
Ioffe S, Szegedy C.
\newblock Batch normalization: Accelerating deep network training by reducing
  internal covariate shift.
\newblock In: ICML; 2015. p. 448--456.

\bibitem{glorot2010understanding}
Glorot X, Bengio Y.
\newblock Understanding the difficulty of training deep feedforward neural
  networks.
\newblock In: AISTATS; 2010. p. 249--256.

\bibitem{kingma2014adam}
Kingma DP, Ba J.
\newblock Adam: A method for stochastic optimization.
\newblock arXiv preprint arXiv:14126980. 2014;.

\bibitem{zeiler2012adadelta}
Zeiler MD.
\newblock Adadelta: An adaptive learning rate method.
\newblock arXiv preprint arXiv:12125701. 2012;.

\bibitem{Hochreiter_1997}
Hochreiter S, Schmidhuber J.
\newblock Long short-term memory.
\newblock Neural Computation. 1997;9(8):1735--1780.
\newblock doi:{10.1162/neco.1997.9.8.1735}.

\bibitem{reimers2017optimal}
Reimers N, Gurevych I.
\newblock Optimal hyperparameters for deep LSTM-networks for sequence labeling
  tasks.
\newblock arXiv preprint arXiv:170706799. 2017;.

\bibitem{dozat2016incorporating}
Dozat T.
\newblock Incorporating Nesterov momentum into Adam.
\newblock In: ICLR Workshops; 2016.

\bibitem{Schuster_1997}
Schuster M, Paliwal KK.
\newblock Bidirectional recurrent neural networks.
\newblock {IEEE} Trans Sig Process. 1997;45(11):2673--2681.
\newblock doi:{10.1109/78.650093}.

\bibitem{Karim_2018}
Karim F, Majumdar S, Darabi H, Chen S.
\newblock {LSTM} fully convolutional networks for time series classification.
\newblock {IEEE} Access. 2018;6:1662--1669.
\newblock doi:{10.1109/access.2017.2779939}.

\bibitem{Karim_2019}
Karim F, Majumdar S, Darabi H.
\newblock Insights into {LSTM} fully convolutional networks for time series
  classification.
\newblock {IEEE} Access. 2019;7:67718--67725.
\newblock doi:{10.1109/access.2019.2916828}.

\bibitem{mukaka2012guide}
Mukaka, Mavuto M.
\newblock A guide to appropriate use of correlation coefficient in medical research
\newblock Malawi Med. J. 2012;24(3):69--71.

\bibitem{Ortigosa_Hern_ndez_2017}
Ortigosa-Hernandez J, Inza I, Lozano JA.
\newblock Measuring the class-imbalance extent of multi-class problems.
\newblock Pattern Recogn. Letters. 2017;98:32--38.
\newblock doi:{10.1016/j.patrec.2017.08.002}.

\end{thebibliography}

\end{document}